\documentclass[a4paper,fleqn]{cas-dc}

\usepackage[numbers]{natbib}
\usepackage{bm}
\usepackage{bigstrut}
\usepackage[numbers]{natbib}
\usepackage{url}
\usepackage{multicol}
\usepackage{graphicx}
\usepackage{algorithm}
\usepackage{algorithmic}

\def\tsc#1{\csdef{#1}{\textsc{\lowercase{#1}}\xspace}}
\tsc{WGM}
\tsc{QE}
\tsc{EP}
\tsc{PMS}
\tsc{BEC}
\tsc{DE}

\begin{document}
\let\WriteBookmarks\relax
\def\floatpagepagefraction{1}
\def\textpagefraction{.001}
\shorttitle{Adaptive Quantile Low-Rank Matrix Factorization}
\shortauthors{Xu et~al.}

\title [mode = title]{Adaptive Quantile Low-Rank Matrix Factorization}                      

\author[1]{Shuang Xu}[orcid= 0000-0003-3576-6914 ,style=chinese,bioid=1]
\ead{shuangxu@stu.xjtu.edu.cn}
\ead[url]{https://xsxjtu.github.io}

\author[1]{Chunxia Zhang}[style=chinese,bioid=2]
\ead{cxzhang@mail.xjtu.edu.cn}
\cormark[1]

\author[1]{Jiangshe Zhang}[style=chinese,bioid=4]
\ead{jszhang@mail.xjtu.edu.cn}

\address[1]{School of Mathematics and Statistics, Xi'an Jiaotong University, Xi'an 710049, China}
\cortext[cor1]{Corresponding author}

\begin{abstract}
Low-rank matrix factorization (LRMF) has received much popularity owing to its successful applications in both computer vision and data mining. By assuming noise to come from a Gaussian, Laplace or mixture of Gaussian distributions, significant efforts have been made on optimizing the (weighted) $L_1$ or $L_2$-norm loss between an observed matrix and its bilinear factorization. However, the type of noise distribution is generally unknown in real applications and inappropriate assumptions will inevitably deteriorate the behavior of LRMF. On the other hand, real data are often corrupted by skew rather than symmetric noise. To tackle this problem, this paper presents a novel LRMF model called AQ-LRMF by modeling noise with a mixture of asymmetric Laplace distributions. An efficient algorithm based on the expectation-maximization (EM) algorithm is also offered to estimate the parameters involved in AQ-LRMF. The AQ-LRMF model possesses the advantage that it can approximate noise well no matter whether the real noise is symmetric or skew. The core idea of AQ-LRMF lies in solving a weighted $L_1$ problem with weights being learned from data. The experiments conducted on synthetic and real datasets show that AQ-LRMF outperforms several state-of-the-art techniques. Furthermore, AQ-LRMF also has the superiority over the other algorithms in terms of capturing local structural information contained in real images.
\end{abstract}

\begin{highlights}
\item A matrix factorization model is proposed to deal with skew noise.
\item Our model can automatically learn the weight of outliers.
\item Our model can capture local structureal information contained in some real images.
\end{highlights}

\begin{keywords}
Low-rank matrix factorization \sep Mixture of asymmetric Laplace distributions \sep  Expectation maximization algorithm \sep Skew noise
\end{keywords}

\maketitle

\section{Introduction}
\label{sec:intro}

Researchers from machine learning \cite{Ye2005Generalized}, computer vision \cite{Udell2014Generalized} and statistics \cite{Koltchinskii2011NUCLEAR} have paid increasing attention to low-rank matrix factorization (LRMF) \cite{Lee1999Learning}. Generally speaking, many real-world modeling tasks can be attributed as the problems of LRMF. The tasks include but are not limited to recommender systems \cite{WU201846}, subspace learning \cite{TOLIC201840,XIONG2019464,DBLP:conf/ijcai/DengLLHTG15,DBLP:journals/pr/YangDN19}, link prediction \cite{WANG2017104}, computational biology \cite{DBLP:conf/ijcai/XuDGSH17,WANG2018410,wang2017transcriptomic} and image denoising \cite{Fei2017Denoising,DBLP:journals/tnn/DengXZTGL16}.

The key idea of LRMF is to approximate a given matrix by the product of two low-rank matrices. Specifically, given an observed matrix $\textbf{X}\in \mathbb{R}^{m\times n}$, LRMF aims at solving the optimization problem
\begin{equation}\label{eq:1}
\min_{\textbf{U},\textbf{V}} ||\bm{\Omega}\odot(\textbf{X}-\textbf{UV}^{\rm T})||,
\end{equation}
where $\textbf{U}\in \mathbb{R}^{m\times r},\textbf{V}\in \mathbb{R}^{n\times r}$ are two low-rank matrices (usually, $r\ll \min(m,n)$) and $\odot$ denotes the Hadamard product, that is, the element-wise product. The indicator matrix $\bm{\Omega}=(\omega_{ij})_{m\times n}$ implies whether some elements are missing, where $\omega_{ij}=1$ if $x_{ij}$ is non-missing and 0 otherwise. The symbol $||\cdot||$ indicates a certain norm of a matrix, in which the most prevalent one is $L_2$ norm. It is well-known that singular value decomposition provides a closed-form solution for $L_2$-norm LRMF without missing entries. With respect to the problems with $\mathbf{X}$ containing missing entries, researchers have presented many fast algorithms such as damped Newton algorithm \cite{DN}, Chen's method \cite{Chen}, and Damped Wiberg (DW) \cite{Okatani2011EfficientAF} to solve Eq. (\ref{eq:1}). In the literature of LRMF, the most popular algorithm is DW proposed in \cite{Okatani2011EfficientAF}. The key idea of DW is to incorporate a damping factor into the Wiberg method to solve the corresponding problem. Although the $L_2$-norm LRMF greatly facilitates theoretical analysis, it provides the best solution in the sense of maximum likelihood principle only when noise is indeed sampled from a Gaussian distribution. If noise is from a heavy-tailed distribution or data are corrupted by outliers, however, $L_2$-norm LRMF may break down. Thereafter, $L_1$-norm LRMF begins to gain increasing interests of both theoretical researchers and practitioners due to its robustness \cite{Ke:2005:RLN:1068507.1068989}. In fact, $L_1$-norm LRMF hypothesizes that noise is from a Laplace distribution. As is often the case with $L_2$-norm LRMF, $L_1$-norm LRMF may provide unexpected results as well if its assumptions are violated.

Because the noise in real data generally deviates far away from a Gaussian or Laplace distribution, analysts are no longer satisfied with $L_1$- or $L_2$-norm LRMF. To further improve the robustness of LRMF, researchers attempt to directly model unknown noise via a mixture of Gaussians (MoG) due to its good property to universally approximate any continuous distribution \cite{Meng2014Robust,Maz1996On}. Nevertheless, the technique cannot fit real noise precisely in some complex cases. For example, in theory, infinite Gaussian components are required to approximate a Laplace distribution. In practice, we only utilize finite Gaussian components due to the characteristics of MoG. On the other hand, Gaussian, Laplace and MoG distributions are all symmetric. In the situations with real noise being skew, it is obviously inappropriate to assume a symmetric noise distribution.

\begin{figure}
	\centering
	\includegraphics[width=1\linewidth]{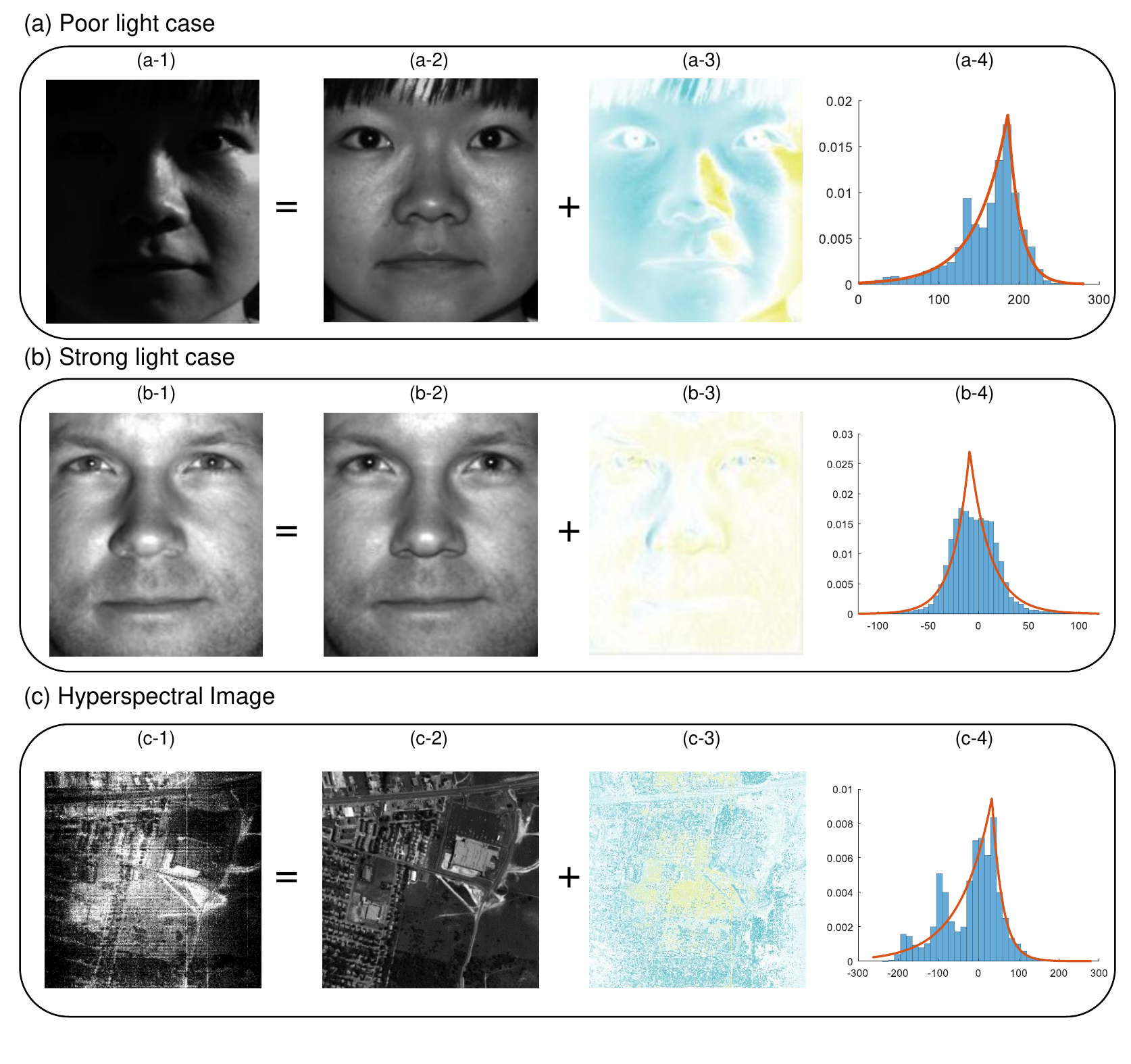}
	\caption{(a) and (b) illustrate two face images corresponding to underexposure and overexposure cases, respectively. In particular, (a-1) and (b-1) are face images captured with improper light sources while (a-2) and (b-2) are face images obtained with proper light sources. (a-3) and (b-3) are residual images in which the yellow (blue) locations indicate positive (negative) values. (a-4) and (b-4) illustrate the histograms of the residual images as well as the PDF curves fitted by ALD with $\alpha_a=115,\kappa_a=0.71,\lambda_a=0.05$ and $\alpha_b=-9,\kappa_b=0.44,\lambda_b=0.11$, respectively. The skewness of the residual face in (a-3) is $-0.72$ whilst that for (b-3) is 0.69. (c) shows a hyperspectral image. Similar to cases (a) and (b), the images from (c-1) to (c-4) are original, de-noised, noise images and the histogram of residuals, respectively. The skewness of the noise image (c-3) is $-0.55$. In (c-4), the fitted ALD is obtained with $\alpha=33$, $\kappa=0.75$ and $\lambda=0.05$. Obviously, the distributions of noise shown here are all asymmetric.}
	\label{fig:fig1}
\end{figure}

As a matter of fact, there are no strictly symmetric noise in real images. For instance, Figure \ref{fig:fig1} illustrates several examples in which the real noise is either skewed to the left (e.g., (a-4) and (c-4)) or the right (e.g., (b-4)). In these situations, the symmetric distributions like Gaussian or Laplace are inadequate to approximate the noise. In statistics, scholars usually make use of quantile regression to deal with an asymmetric noise distribution \cite{Davino2014Quantile_Reg}. Consider a simple case that there is only one covariate $X$, the quantile regression coefficient $\beta$ can be obtained by
\begin{equation}
\label{eq:quan_loss}
\hat{\beta}_{\kappa}={\underset {\beta}{\mbox{arg min}}}\sum _{i=1}^{n}\rho _{\kappa}(y_{i}-x_{i}\beta),
\end{equation}
where $\{(y_i,x_i)\}_{i=1}^n$ are $n$ observations and $\kappa$ is a pre-defined asymmetry parameter. Moreover, the quantile loss $\rho_{\kappa}(\cdot)$ is defined as
\begin{equation}
\label{eq:quan_loss_fun}
\begin{aligned}
\rho_{\kappa}(\epsilon)&=\epsilon\left[\kappa-\mathbb{I}(\epsilon<0)\right]\\
&=|\epsilon| [\kappa \mathbb{I}(\epsilon\geq0)+ (1-\kappa)\mathbb{I}(\epsilon<0)]\\
\end{aligned}
\end{equation}  
with $\mathbb{I}(\cdot)$ being the indicator function. Evidently, the quantile loss with $\kappa=1/2$ corresponds to the $L_1$-norm loss. From the Bayesian viewpoint, the estimate obtained by minimizing the quantile loss in (\ref{eq:quan_loss}) coincides with the result by assuming noise coming from an asymmetric Laplace distribution (ALD) \cite{Kozumi2011Gibbs,Keming2005A}.

To overcome the shortcomings of existing LRMF methods that they assume a specific type of noise distribution, we present in this paper an adaptive quantile LRMF (AQ-LRMF) algorithm. The key idea of AQ-LRMF is to model noise via a mixture of asymmetric Laplace distributions (MoAL). Due to the existence of some latent variables, the expectation maximization (EM) algorithm is employed to estimate the parameters in AQ-LRMF under the maximum likelihood framework. The novelty of AQ-LRMF and our main contributions can be summarized as follows.
\begin{enumerate}
	\item[(1).] The M-step of the EM algorithm corresponds to a weighted $L_1$-norm LRMF, where the weights encode the information about skewness and outliers. 	
	\item[(2).] The weights are automatically learned from data under the framework of EM algorithm.
	\item[(3).] Different from quantile regression, our method does not need to pre-define the asymmetry parameter of quantile loss, because it is adaptively determined by data.
	\item[(4).] Our model can capture local structural information contained in some real images, although we do not encode it into our model.
\end{enumerate}
Our conducted experiments show that AQ-LRMF can effectively approximate many different kinds of noise. If the noise has a strong tendency to take a particular sign, AQ-LRMF will produce better estimates than a method which assumes a symmetric noise distribution. In comparison with several state-of-the-art methods, the superiority of our method is demonstrated in both synthetic and real-data experiments such as image inpainting, face modeling, hyperspectral image (HSI) construction and so on. The code of this paper is available at \url{https://xsxjtu.github.io/Projects/MoAL/main.html}.

The rest of the paper is organized as follows. Section \ref{sec:related_work} presents related work of LRMF. In section \ref{sec:AQ-LRMF}, we propose the AQ-LRMF model and also provide an efficient learning algorithm for it. Section \ref{sec:experiments} includes experimental studies. At last, some conclusions are drawn in section \ref{sec:conclusions}.

\section{Related work}
\label{sec:related_work}

The study of robust LRMF has a long history. Srebro and Jaakkola \cite{icml2003SrebroJ03} suggested to use a weighted $L_2$ loss to improve LRMF's robustness to noise and missing data. The problem can be solved by a simple but efficient EM algorithm. However, its capability strongly relies on the chosen weights while it is not easy to automatically select proper weights. Since then, the research community began to replace $L_2$ loss with $L_1$ loss. One of the earliest explorations is made by Ke and Kanade \cite{Ke:2005:RLN:1068507.1068989}. They solved the $L_1$-norm LRMF by alternated linear or quadratic programming, but the speed is slow. Thereafter, many researchers attempted to develop some variants of $L_1$-norm LRMF to enhance its running speed as well as performance. Roughly speaking, the improved $L_1$-norm LRMF can be classified into two groups.

On the one hand, researchers strived to propose fast numerical algorithms for $L_1$-norm LRMF. Under this framework, Eriksson and Hengel \cite{L1Wiberg} developed the $L_1$-Wiberg algorithm for calculating the low-rank factorization of a matrix which minimizes the $L_1$ norm in the presence of missing data. Meng {\it et al.} \cite{CWM} proposed a computationally efficient algorithm, cyclic weighted median (CWM) method, by solving a sequence of scalar minimization sub-problems to obtain the optimal solution. Recently, Kim {\it et al.} \cite{Kim2015Efficient} used alternating rectified gradient method to solve a large-scale $L_1$-norm LRMF.

On the other hand, researchers tried to improve $L_1$-norm LRMF's performance by inserting a penalty into the objective function. Okutomi {\it et al.} \cite{RegL1ALM} modified the objective function of $L_1$-Wiberg by adding the nuclear norm of $\textbf{V}$ and the orthogonality constraint on $\textbf{U}$. This method has been shown to be effective in addressing structure from motion issue. Inspired by majorization-minimization technique, Lin {\it et al.} \cite{Lin2017Robust} proposed LRMF-MM to solve an LRMF optimization task with $L_1$ loss plus the $L_2$-norm penalty that is placed on $\mathbf{U}$ and $\mathbf{V}$. In each step, they upper bound the original objective function by a strongly convex surrogate and then minimize the surrogate. Li {\it et al.} \cite{Li2017Efficient} considered a similar problem, but they replace the $L_2$-norm penalty imposed on $\mathbf{U}$ with $\mathbf{U}^{\rm T}\mathbf{U}=\mathbf{I}$. This model is solved by augmented Lagrange multiplier method. Furthermore, the authors of \cite{Li2017Efficient} designed a heuristic rank estimator for their model. Even though the above-mentioned approaches improved $L_1$-norm LRMF from a certain aspect, one has to notice that $L_1$ loss actually corresponds to the Laplace-distributed noise. Put in another way, these methods implicitly assume that the noise comes from a Laplace distribution. When the real distribution of noise deviates too far from Laplace, the robustness of $L_1$ LRMF will be suspectable.

Recently, the research community began to focus on probabilistic extensions of robust matrix factorizations. Generally speaking, it is assumed that $\mathbf{X} = \mathbf{UV}^{\rm T}+\mathbf{E}$, where $\mathbf{E}$ is a noise matrix. Lakshminarayanan {\it et al.} \cite{Lakshminarayanan2011Robust} replaced Gaussian noise with Gaussian scale mixture noise. Nevertheless, it may be ineffective when processing heavy-tailed (such as Laplace-type) noise. Wang {\it et al.} \cite{Wang2012} proposed a probabilistic $L_1$-norm LRMF, but they did not employ a fully Bayesian inference process.
Beyond Laplace noise, Meng and Torre \cite{Meng2014Robust} presented a robust LRMF with unknown noise modeled by an MoG. In essence, the method iteratively optimizes $
\min_{\textbf{U},\textbf{V},\boldsymbol{\theta}} ||\mathbf{W}(\boldsymbol{\theta})\odot(\textbf{X}-\textbf{UV}^{\rm T})||_{L_2}$, where $\boldsymbol{\theta}$ are the MoG parameters which are automatically updated during optimization, and $\mathbf{W}(\boldsymbol{\theta})$ is the weight function of $\boldsymbol{\theta}$. Due to the benefit to adaptively assign small weights to corrupted entries, MoG-LRMF has been reported to be fairly effective. More recently, Cao {\it et al.} \cite{Cao2015Low} presented a novel LRMF model by assuming noise as a mixture of exponential power (MoEP) distributions and offered both a generalized expectation maximization (GEM) algorithm and a variational GEM to infer all parameters involved in their proposed model.

In addition, it is worth mentioning that robust principle component analysis (robust PCA) \cite{Candes2011Robust} considers an issue similar to LRMF, that is,
\begin{equation}\label{eq:rpca}
\min_{\textbf{A},\textbf{E}} {\rm rank}(\mathbf{A})+\lambda ||\mathbf{E}||_{L_0} \quad {\rm s.t.} \ \mathbf{X} =\mathbf{A}+ \mathbf{E}.
\end{equation}
The underlying assumption of robust PCA is that the original data can be decomposed into the sum of a low-rank matrix and a sparse outlier matrix (i.e., the number of non-zero elements in $\mathbf{E}$ is small). Clearly, $\mathbf{A}$ plays the same role as the product of $\mathbf{U}$ and $\mathbf{V}^{\rm T}$. Since Eq. (\ref{eq:rpca}) involves a non-convex objective function, \cite{Candes2011Robust} consider a tractable convex alternative, called principal component pursuit, to handle the corresponding problem, namely,
\begin{equation}\label{eq:pcp}
\min_{\textbf{A},\textbf{E}} ||\mathbf{A}||_*+\lambda ||\mathbf{E}||_{L_1} \quad {\rm s.t.} \ \mathbf{X} =\mathbf{A}+ \mathbf{E},
\end{equation}
where $||\cdot||_*$ denotes the nuclear norm. Nevertheless, principal component pursuit may sometimes fail to recover $\mathbf{E}$ when the real observation is also corrupted by a dense inlier matrix. To overcome this shortcoming, Zhou {\it et al.} \cite{Zhou2010Stable} proposed the stable principal component pursuit (SPCP) by solving \begin{equation}\label{eq:spcp}
\min_{\textbf{A},\textbf{E}} ||\mathbf{A}||_*+\lambda ||\mathbf{E}||_{L_1} \quad {\rm s.t.} \ ||\mathbf{X}-\mathbf{A}- \mathbf{E}||_{L_2}\le \varepsilon.
\end{equation}
Actually, the underlying assumption of SPCP is $\mathbf{X}=\mathbf{A}+\mathbf{N}+\mathbf{E}$, where $\mathbf{A}$ is a low-rank component, $\mathbf{E}$ is a sparse matrix representing the gross sparse errors (i.e., outliers) in the observed data $\mathbf{X}$ and $\mathbf{N}$ is the small-magnitude noise that can be modeled by a Gaussian distribution. Both theoretical analysis and experiments have shown that SPCP guarantees the stable recovery of $\mathbf{E}$ \cite{Zhou2010Stable,Candes2011Robust}.

Actually, our model AQ-LRMF (details are provided in section \ref{sec:AQ-LRMF}) is a probabilistic extension of robust matrix factorization. The differences between AQ-LRMF and existing approaches can be summarized as follows. First, the noise in AQ-LRMF is assumed to be asymmetric (i.e., the noise is modeled with an MoAL), while the noise in existing methods is governed by a symmetric distribution, such as Gaussian and Laplacian. Second, the parameters in AQ-LRMF are inferred by an EM algorithm. In the M-step, the optimization with regard to $\mathbf{U}$ and $\mathbf{V}$ is cast into a weighted $L_1$-norm LRMF, where the weights are automatically learned from data. Meanwhile, the weights embody the information about outliers and skewness. In contrast, the M-step in MoG-LRMF leads to a weighted $L_2$-norm LRMF. Because $L_1$ norm is more robust to noise and outliers, AQ-LRMF also inherits this good property to perform better than MoG-LRMF in handling various kinds of noise.

\section{Adaptive Quantile LRMF (AQ-LRMF)}
\label{sec:AQ-LRMF}

\subsection{Motivation}
\label{sec:motivation}

Generally speaking, researchers employ the $L_2$ or $L_1$ loss function when solving a low-rank matrix factorization problem. As argued in introduction, $L_2$ or $L_1$ loss implicitly hypothesizes that the noise distribution is symmetric. Nevertheless, the noise in real data is often asymmetric and Fig. \ref{fig:fig1} illustrates several examples.

In Fig. \ref{fig:fig1}, there are two face images and a hyperspectral image. Fig. \ref{fig:fig1} (a) displays a face image that is captured with a poor light source. There are cast shadows in a large area, while there exists an overexposure phenomenon in a small area. As a result, the noise is negative skew. By contrast, Fig. \ref{fig:fig1} (b) illustrates a face image which is captured under a strong light source. Because of the camera range settings, there are saturated pixels, especially on the forehead. Under this circumstance, the noise is positive skew. Fig. \ref{fig:fig1} (c) shows a hyperspectral image that is mainly corrupted by stripe and Gaussian noise. Its residual image indicates that the signs of the noise are unbalanced, i.e., more pixels are corrupted by noise with negative values. Actually, the skewness values of three residual (noise) images are $-0.72$, 0.69 and $-0.55$, respectively. Note that a symmetric distribution has skewness 0, the noise contained in these real data sets is thus asymmetric.

As a matter of fact, the noise in real data can hardly be governed by a strictly symmetric probability distribution. Therefore, it is natural to utilize an asymmetric distribution to model realistic noise. In statistics, researchers usually make use of a quantile loss function defined in (\ref{eq:quan_loss_fun}) to address this issue. It has been shown that quantile loss function corresponds to the situation that noise is from an asymmetric Laplace distribution \cite{Kozumi2011Gibbs,Keming2005A}. In order to further improve the performance of LRMF, we attempt to use a mixture of asymmetric Laplacian distributions (MoAL) to approximate noise.

\subsection{Asymmetric Laplace distribution}
\label{sec:ALD}

\begin{figure}
	\centering
	\includegraphics[width=\linewidth]{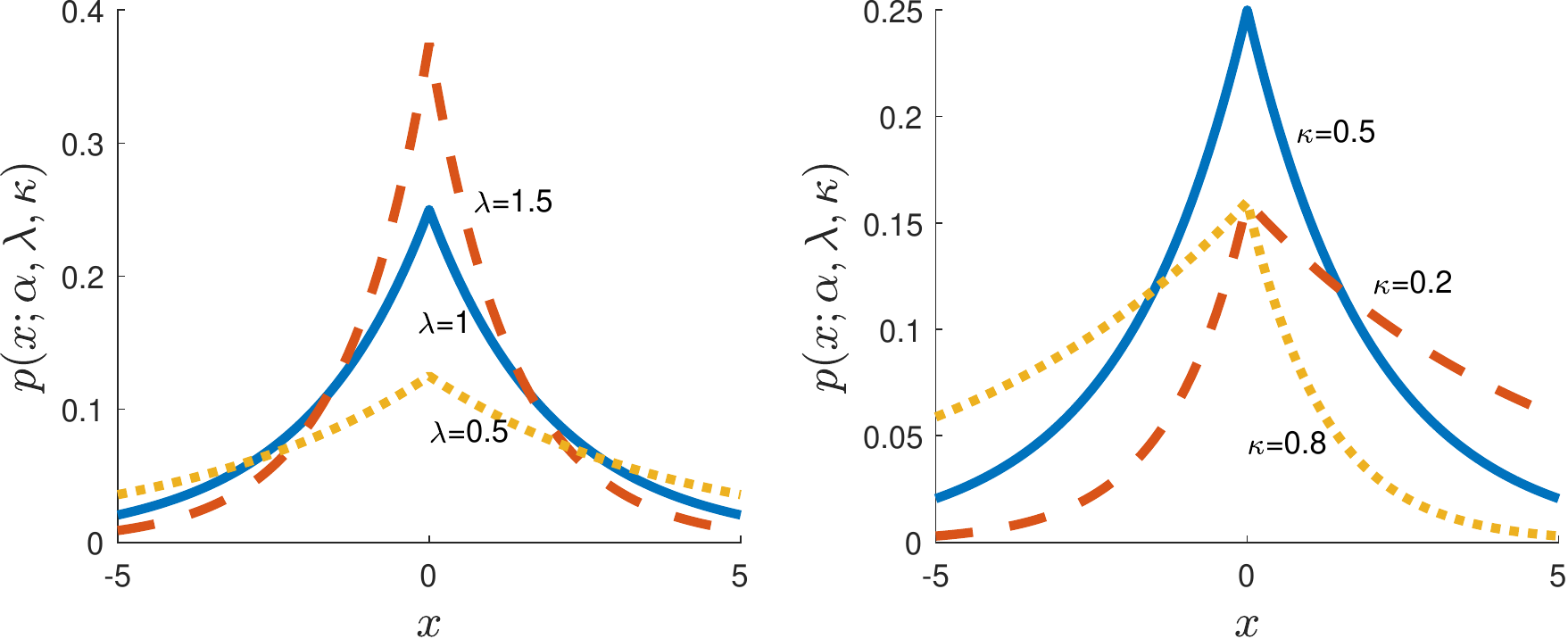}
	\caption{The PDF curves of ALDs. The location parameter is $\alpha=0$. Left: $\kappa=0.5$; right: $\lambda=1$.}
	\label{fig:ald3parameters}
\end{figure}

In what follows, we use $AL(\epsilon|\alpha, \lambda, \kappa)$ to denote an ALD with location, scale and asymmetric parameters $\alpha$, $\lambda>0$ and $0<\kappa<1$, respectively. Its probability distribution function (PDF) \cite{Keming2005A} is
\begin{equation}
\begin{aligned}
&p(x;\alpha,\lambda ,\kappa)\\
=& \lambda \kappa(1-\kappa)
{\begin{cases}
	\exp \left( \lambda(1-\kappa)(x-\alpha)\right), & {\text{if}}\quad x<\alpha;\\
	\exp\left( -\lambda \kappa (x-\alpha)\right),                & {\text{if}}\quad x\geq \alpha;
	\end{cases}}\\
=& \lambda \kappa(1-\kappa) \exp\left( -|x-\alpha|\lambda[\kappa \mathbb{I}(x-\alpha\geq0) \right.\\
& \left. + (1-\kappa)\mathbb{I}(x-\alpha<0)]\right).
\end{aligned}
\end{equation}
Obviously, the location parameter $\alpha$ is exactly the mode of an ALD. In Fig. \ref{fig:ald3parameters}, we demonstrate the PDF curves for several ALDs with different parameters. In general, the skewness of an ALD, say, $\mathrm{sk_{ALD}}$, takes value in the interval $(-2,2)$ and it is controlled by the asymmetry parameter $\kappa$. An ALD is positive skew if $0<\kappa<0.5$, and is negative skew if $0.5<\kappa<1$. If $\kappa=0.5$, the ALD becomes a Laplace distribution. The smaller the scale parameter $\lambda$ is, the more heavy-tailed an ALD is.

It is worthwhile that skew Gaussian distributions \cite{azzalini1996the} are also prevailing in both theory and applications. However, it is not ideal for the analysis of LRMF. On the one hand, the PDF of a skew Gaussian distribution is complex. On the other hand, its skewness lies in $(-1,1)$ which is only a subset of the range of $\mathrm{sk_{ALD}}$. Due to this fact, the fitting capability of an ALD is greater than that of a skew Gaussian distribution.

\subsection{AQ-LRMF model}
\label{sec:AQ-LRMF_detail}

To enhance the robustness of LRMF in situations with skew and heavy-tailed noise, we propose an adaptive quantile LRMF (AQ-LRMF) by modeling unknown noise as an MoAL. In particular, we consider a generative model of the observed matrix $\textbf{X}\in\mathbb{R}^{m\times n}$. For each entry $x_{ij}$, suppose that there is
\begin{equation}
x_{ij} = \textbf{u}_i\textbf{v}_j^{\rm T} + \epsilon_{ij},
\end{equation}
where $\textbf{u}_i$ is the $i$th row of $\textbf{U}$, $\textbf{v}_j$ is the $j$th row of $\textbf{V}$, and $\epsilon_{ij}$ is the noise. In AQ-LRMF, we assume that $\epsilon_{ij}$ is distributed as an MoAL, namely,
\begin{equation}
p(\epsilon_{ij}) = \sum_{s=1}^{S} \pi_s AL_s(\epsilon_{ij}| 0, \lambda_s, \kappa_s),
\end{equation}
in which $AL_s(\epsilon_{ij}|0, \lambda_s, \kappa_s)$ stands for an asymmetric distribution with parameters $\alpha=0, \lambda = \lambda_s$ and $\kappa=\kappa_s$. Meanwhile, $\pi_s$ indicates the mixing proportion with $\pi_s\geq 0$ and $\sum_{s=1}^S\pi_s=1$, and $S$ means the number of mixture components.

To facilitate the estimation of unknown parameters, we introduce some latent binary variables $z_{ij1}, z_{ij2}, \cdots$, $z_{ijS}$ where $z_{ijs}\in\{0,1\}$ and $\sum_{s=1}^Sz_{ijs}=1$. To ease presentation, let each noise $\epsilon_{ij}$ be equipped with an indicator vector $\textbf{z}_{ij} = (z_{ij1}, z_{ij2}, \cdots, z_{ijS})^{\rm T}$. Here, $z_{ijs}=1$ indicates that the noise $\epsilon_{ij}$ is drawn from the $s$th AL distribution. Evidently, $\textbf{z}_{ij}$ follows a multinomial distribution, i.e., $\textbf{z}_{ij}\sim \mathcal{M}(\pi_1,\cdots,\pi_S)$. Under these assumptions, we can have
\begin{equation}
p(\epsilon_{ij}) = \prod_{s=1}^{S} \left[\pi_s AL_s(\epsilon_{ij}| 0, \lambda_s, \kappa_s)\right] ^{z_{ijs}}.
\end{equation}

Now, it is easy to obtain the probability of $x_{ij}$ as
\begin{equation}
p(x_{ij}| \textbf{u}_i, \textbf{v}_j, \bm{\lambda}, \textbf{K}, \bm{\pi} ) =
\prod_{s=1}^{S} \left[\pi_s AL_s(x_{ij}| \textbf{u}_i\textbf{v}_j^{\rm T}, \lambda_s, \kappa_s) \right] ^{z_{ijs}},
\end{equation}
where $\bm{\lambda} = \{\lambda_1,\lambda_2,\cdots,\lambda_S\}$, $\mathbf{K} = \{\kappa_1,\kappa_2,\cdots,\kappa_S\}$ and $\bm{\pi} = \{\pi_1,\pi_2,$ $\cdots,\pi_S\}$ are unknown parameters. To estimate $\textbf{U},\textbf{V}$ as well as $\bm{\lambda}, \textbf{K}, \bm{\pi}$, we employ the maximum likelihood principle. Consequently, the goal is to maximize the log-likelihood function of complete data shown below, namely,
\begin{equation}\label{eq12}
\begin{aligned}
&\ell(\textbf{U},\textbf{V},\bm{\lambda}, \textbf{K}, \bm{\pi}) \\
=&
\sum_{(i,j)\in\Omega}\sum_{s=1}^{S} z_{ijs} \left[ \log AL_s(x_{ij}| \textbf{u}_i\textbf{v}_j^{\rm T}, \lambda_s, \kappa_s)+\log\pi_s\right] ,
\end{aligned}
\end{equation}
where $\Omega$ denotes the index set of the non-missing entries of data. Subsequently, we will discuss how to maximize the log-likelihood function $\ell(\textbf{U},\textbf{V},\bm{\lambda}, \textbf{K}, \bm{\pi})$ to get our interested items. 

\subsection{Learning of AQ-LRMF}
\label{sec:learning}

Since each $x_{ij}$ associates with an indicator vector $\textbf{z}_{ij} = (z_{ij1}, z_{ij2}, \cdots, z_{ijS})^{\rm T}$ in which $z_{ijk}$'s $(k=1,\cdots,S)$ are latent variables, the EM algorithm \cite{em} is utilized to train the AQ-LRMF model. Particularly, the algorithm needs to iteratively implement the following two steps (i.e., E-step and M-step) to maximize the likelihood of the corresponding problem until the algorithm converges. For ease of exposition, we let $e_{ij}=x_{ij} - \textbf{u}_i\textbf{v}_j^{\rm T}$ and abbreviate $AL_s(e_{ij}|0, \lambda_s, \kappa_s)$ as $AL_s(e_{ij})$ in the following discussions.

\noindent\textbf{E-step}: Compute the conditional expectation of the latent variable $z_{ijs}$ as
\begin{equation}\label{E-step}
\gamma_{ijs} = E(z_{ijs}|x_{ij}) =
\frac{\pi_s AL_s(e_{ij})}
{\sum_{a=1}^{S}\pi_a AL_a(e_{ij})}.
\end{equation}

In order to attain the updating rules of other parameters, we need to compute the $ Q $-function. According to the working mechanism of EM algorithm, the $ Q $-function can be obtained by taking expectation of the log-likelihood function shown in (\ref{eq12}) with regard to the conditional distribution of the latent variables $z_{ij1},z_{ij2},\cdots,z_{ijS}$. Specifically, it can be derived as
\begin{equation}
\begin{aligned}
Q &= E_{\mathbf{Z}\mid \mathbf{X}} [\ell(\textbf{U},\textbf{V},\bm{\lambda}, \textbf{K}, \bm{\pi})]\\
&= E_{\mathbf{Z}\mid \mathbf{X}} \{\sum_{(i,j)\in\Omega}\sum_{s=1}^{S} z_{ijs} \left[ \log AL_s(e_{ij}| 0, \lambda_s, \kappa_s)+\log\pi_s\right]\} \\
&=\sum_{(i,j)\in\Omega}\sum_{s=1}^{S} \gamma_{ijs} \left[ \log AL_s(e_{ij}| 0, \lambda_s, \kappa_s)+\log\pi_s\right] \\
&=\sum_{(i,j)\in\Omega}\sum_{s=1}^{S} \gamma_{ijs} \{  \log\pi_s +\log \lambda_s \kappa_s(1-\kappa_s) \\
&\quad - |e_{ij}|\lambda_s\left[ (1-\kappa_s)\mathbb{I}(e_{ij}<0)+\kappa_s\mathbb{I}(e_{ij}\geq0)\right] \}\\
&\equiv \sum_{(i,j)\in\Omega}\sum_{s=1}^S \gamma_{ijs}\left[ \log\kappa_s(1-\kappa_s)\lambda_s\pi_s-\lambda_s\rho_{ijs}|e_{ij}| \right],
\end{aligned}
\end{equation}
where
\begin{equation}\label{eq:rho}
\rho_{ijs}=\left[ (1-\kappa_s)\mathbb{I}(e_{ij}<0)+\kappa_s\mathbb{I}(e_{ij}\geq0)\right].
\end{equation}

\noindent\textbf{M-step}: Maximize the $Q$-function by iteratively updating its parameters as follows.
\begin{enumerate}
	\item[(1).]{\textbf{Update $\pi_s$}: To attain the update for $\mathbf{\pi_s}$, we need to solve the following constrained optimization problem
		\begin{equation}
		\max_{\pi_s} \sum_{(i,j)\in\Omega}\sum_{s=1}^S \gamma_{ijs} \log\pi_s,\quad {\rm s.t.}\quad	\sum_{s=1}^S \pi_{s} = 1,
		\end{equation}
		via the Lagrangian multiplier method. By some derivations, we have
		\begin{equation}\label{M-step1-1}
		\pi_s = \frac{N_s}{N},\quad {\rm where}\quad N_s = \sum_{(i,j)\in\Omega} \gamma_{ijs},
		\end{equation}
		in which $N$ stands for the cardinality of $\Omega$.}
	\item[(2).]{\textbf{Update $\lambda_s$}: Compute the gradient $\displaystyle\frac{\partial Q}{\partial \lambda_s}$ and let it be zero. Consequently, the update of $\lambda_s$ can be obtained as
		\begin{equation}\label{M-step1-2}
		\lambda_s = \frac{N_s}
		{\sum_{(i,j)\in\Omega} \rho_{ijs}\gamma_{ijs}|e_{ij}|}.
		\end{equation}}
	\item[(3).]{\textbf{Update $\kappa_s$}: Compute the gradient $\displaystyle\frac{\partial Q}{\partial \kappa_s}$ and let it be zero, we can have
		\begin{equation}\label{M-step1-33}
		\begin{aligned}
		\eta_s\kappa_s^2-(2N_s+\eta_s)\kappa_s+N_s = 0,
		\end{aligned}
		\end{equation}
		where the coefficients $\eta_s = \lambda_s\sum_{(i,j)\in\Omega}\gamma_{ijs}e_{ij}$. Evidently, Eq. (\ref{M-step1-33}) is a two-order equation with regard to $\kappa_s$ and it has a unique root satisfying $0<\kappa_s<1$, that is,
		\begin{equation}\label{M-step1-3}
		\kappa_s = \frac{2N_s+\eta_s-\sqrt{4N_s^2+\eta_s^2}}{2\eta_s}.
		\end{equation}}
	\item[(4).]{\textbf{Update $\mathbf{U,V}$}: By omitting some constants, the objective function to optimize $\mathbf{U,V}$ can be rewritten as
		\begin{equation}\label{m-step3}
		\begin{aligned}
		&	\max -\sum_{(i,j)\in\Omega}\sum_{s=1}^S \lambda_s\gamma_{ijs}\rho_{ijs} |x_{ij}-\mathbf{u}_i\mathbf{v}_j^{\rm T}|\\
		\Leftrightarrow & \min \sum_{i=1}^{m}\sum_{j=1}^{n} w_{ij} |x_{ij}-\mathbf{u}_i\mathbf{v}_j^{\rm T}|\\
		\Leftrightarrow & \min ||\textbf{W}\odot(\textbf{X}-\textbf{UV}^{\rm T})||_{L_1},
		\end{aligned}
		\end{equation}
		where the $(i,j)$th entry of $\textbf{W}$ is
		\begin{equation}\label{eq:w_ij}
		w_{ij}=
		\begin{cases}
		\sum_{s=1}^S \lambda_s\gamma_{ijs}\rho_{ijs}, &\ {\rm if}\ (i,j)\in\Omega,\\
		0, &\ {\rm if}\ (i,j)\notin \Omega.
		\end{cases}
		\end{equation}}
\end{enumerate}
Hence, the optimization problem in Eq. (\ref{m-step3}) is equivalent to the weighted $L_1$-LRMF, which can be solved by a fast off-the-shelf algorithm. In this paper, the cyclic weighted median filter (CWM) \cite{CWM} is employed to solve Eq. (\ref{m-step3}) and the detailed derivations will be introduced in the next subsection.

Here, it is interesting that the M-step in AQ-LRMF is the same as that of MoG-LRMF \cite{Meng2014Robust}, except that the latter one minimizes a weighted $L_2$ loss. Due to this feature, AQ-LRMF is more robust than MoG-LRMF. On the other hand, each weight of MoG-LRMF embodies the information about whether the corresponding entry is an outlier. For each weight of AQ-LRMF, it actually contains additional information about the sign of bias. In particular, $\lambda_s$ is the scale parameter and the entries with smaller $\lambda_s$ correspond to outliers. According to the definition of $\rho_{ijs}$ in Eq. (\ref{eq:rho}), we know that $\rho_{ijs}$ is a function of the skewness parameter $\kappa_s$. If the residual $e_{ij}\ge0$, $\rho_{ijs}=\kappa_s$ and $\rho_{ijs}=1-\kappa_s$ otherwise. Hence, the weights assigned to two different points still differ if two residuals with the same absolute value have different signs. In conclusion, AQ-LRMF has more capacity to process heavy-tailed skew data.

\begin{figure}
	\centering
	\includegraphics[width=\linewidth]{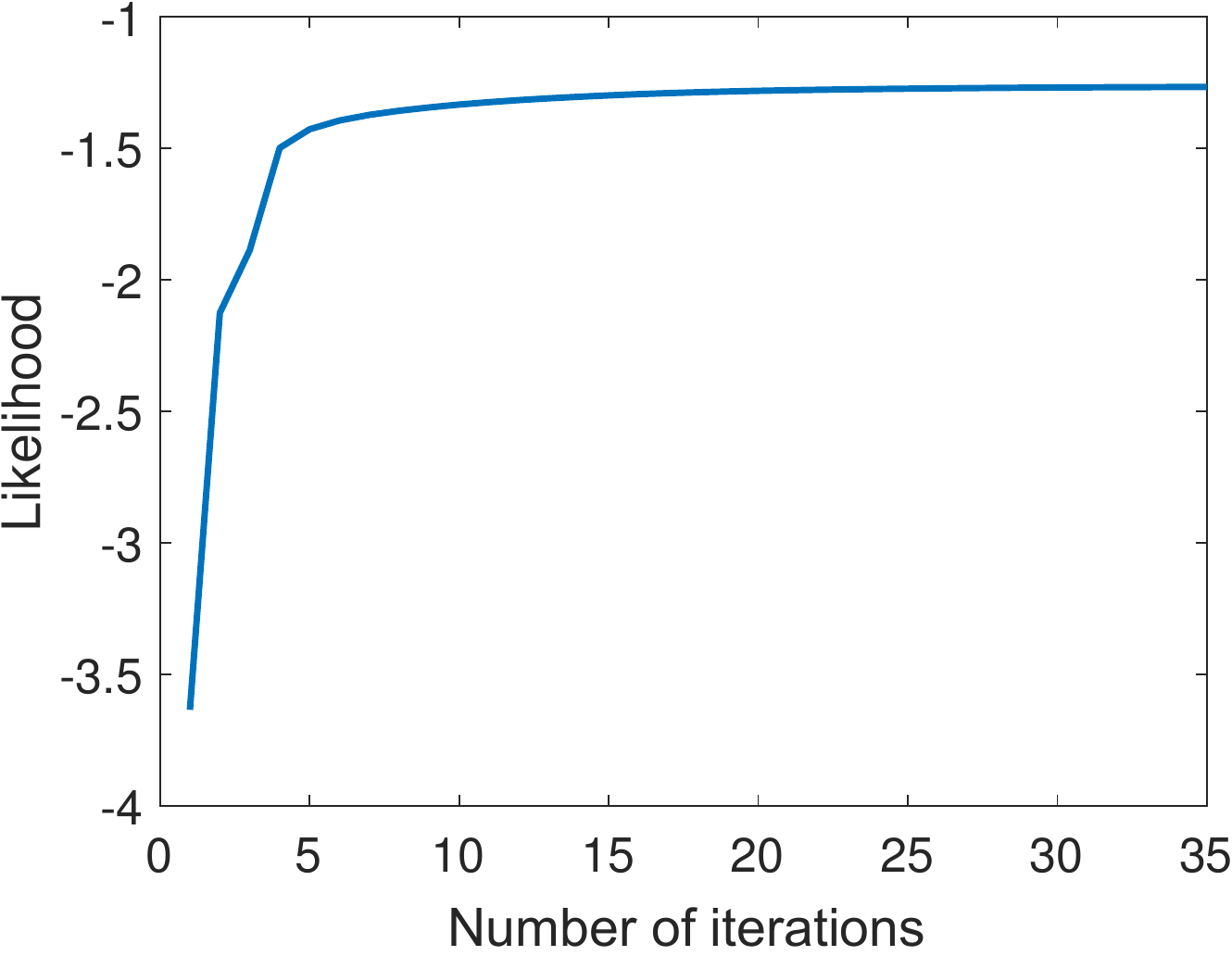}
	\caption{In a synthetic experiment, how the likelihood value varies as the number of iterations increases.}
	\label{fig:llh}
\end{figure}

Based on the above analysis, we summarize the main steps to learn the parameters involved in AQ-LRMF as shown in Algorithm \ref{alg:moal}. We now discuss the computational complexity of Algorithm \ref{alg:moal}. The complexity of updating $\bm{\gamma}$ is $O(mnS)$ and that of updating $\bm{\pi}$ and $\bm{\lambda}$ is the same. As for the complexity to update $\bm{\kappa}$, it is $O(S)$. At last, the complexity to update $\mathbf{U},\mathbf{V}$ will be $O(mnS)$ if Eq. (\ref{m-step3}) is solved by CWM. Thus, the total time complexity of Algorithm \ref{alg:moal} is $O(T(mnS+S))$, where $T$ is the number of iterations for the algorithm to reach convergence. Note that Algorithm \ref{alg:moal} is derived by EM algorithm, it can thus converge to a local optimum within finite iterations since the likelihood does not decrease in each step. As an example, Fig. \ref{fig:llh} depicts how the likelihood value varies as the number of iteration increases in a synthetic experiment (please see the detailed settings in subsection \ref{sec:synthetic_exp}). It is shown that the likelihood value increases quickly in the first few iterations, and then it gradually levels off. Finally, the algorithm converges at the $35$th iteration.
\begin{algorithm}
	\caption{Learning algorithm of AQ-LRMF}
	\label{alg:moal}
	\begin{algorithmic}[1]
		\REQUIRE ~~\\ 
		The observed matrix $\textbf{X}$ of order $m\times n$; the index set $\Omega$ of non-missing entries of $\textbf{X}$; number of components $S$ in MoAL.
		\ENSURE ~~\\ 
		$\textbf{U},\textbf{V}$.
		\STATE Initialize $\textbf{U},\textbf{V},\bm{\lambda}, \textbf{K}, \bm{\pi}$.
		\STATE (Initial E-step): Evaluate $\gamma_{ijs}$ by Eq. (\ref{E-step}), $i=1,...,m;\ j=1,...,n;\ s=1,\cdots,S$.
		\WHILE {the convergence criterion does not satisfy}
		\STATE (M-step 1): Update $\pi_s, \lambda_s, \kappa_s$ ($s=1,\cdots,S$) with Eqs. (\ref{M-step1-1}), (\ref{M-step1-2}) and (\ref{M-step1-3}), respectively.
		\STATE (E-step 1): Evaluate $\gamma_{ijs}$ by Eq. (\ref{E-step}), $i=1,...,m;\ j=1,...,n;\ s=1,\cdots,S$.
		\STATE (M-step 2): Update $\textbf{U},\textbf{V}$ by solving Eq. (\ref{m-step3}) with the CWM method.
		\STATE (E-step 2): Evaluate $\gamma_{ijs}$ by Eq. (\ref{E-step}), $i=1,...,m;\ j=1,...,n;\ s=1,\cdots,S$.
		\STATE (Tune $S$): For each pair $(i,j)\in \Omega$, compute its noise component index $\textbf{C}(i,j)=\arg\max_s \gamma_{ijs}$. Remove any ALD components which are not in $\textbf{C}$. Let $S$ be the current number of ALD components.
		\ENDWHILE
	\end{algorithmic}
\end{algorithm}

\subsection{Solution of the weighted $L_1$-LRMF}
\label{sec:solution}

As stated in the last subsection, the learning of AQ-LRMF can be cast into a weighted $L_1$-LRMF problem. Now we will provide more details about how to solve it (i.e., how to update $\textbf{U},\textbf{V}$ by Eq. (\ref{m-step3})) with the CWM method \cite{CWM}.

Essentially, CWM minimizes the objective via solving a series of scalar minimization subproblems. Let $\mathbf{U}=(\tilde{\mathbf{u}}_1,\tilde{\mathbf{u}}_2,\cdots,\tilde{\mathbf{u}}_r)\in\mathbb{R}^{m\times r}$ and $\mathbf{V}=(\tilde{\mathbf{v}}_1,\tilde{\mathbf{v}}_2,\cdots,\tilde{\mathbf{v}}_r)\in\mathbb{R}^{n\times r}$, respectively. To update $v_{ji}\ (j=1,\cdots,n;i = 1,\cdots,r)$, we assume that the other parameters have been estimated. As a result, the original problem can be rewritten as the optimization problem regarding $v_{ji}$, i.e.,
\begin{equation}\label{eq:CWM}
\begin{aligned}
||\textbf{W}\odot(\textbf{X}-\textbf{UV}^T)||_{L_1}
= ||\textbf{W}\odot(\textbf{X}-\sum_{j=1}^{r}\tilde{\mathbf{u}}_j\tilde{\mathbf{v}}_j^T)||_{L_1}\\
= ||\textbf{W}\odot(\textbf{E}_i-\tilde{\mathbf{u}}_i \tilde{\mathbf{v}}_i^T)||_{L_1}
= ||\tilde{\textbf{w}}_j\odot(\tilde{\textbf{e}}_j^i-\tilde{\mathbf{u}}_i v_{ji})||_{L_1} + \text{c},\\
\end{aligned}
\end{equation}
where $\mathbf{E}_i = \mathbf{X}-\sum_{j\ne i}\tilde{\mathbf{u}}_j\tilde{\mathbf{v}}_j^T$, and $\tilde{\textbf{w}}_j$ and $\tilde{\textbf{e}}_j^i$ are $j$th column of $\mathbf{W}$ and $\mathbf{E}_i$, respectively. In Eq. (\ref{eq:CWM}), $\text{c}$ denotes a constant term that does not depend on $v_{ji}$. In this way, the optimal $v_{ji}$, say $v_{ji}^*$, can be easily attained by the weighted median filter. Specifically, let $\mathbf{e}=\tilde{\textbf{w}}_j\odot\tilde{\textbf{e}}_j^i$ and $\mathbf{u}=\tilde{\textbf{w}}_j\odot\tilde{\mathbf{u}}_i$, we can reformulate Eq. (\ref{eq:CWM}) as
\begin{equation} \label{eq:reform_CWM}
\begin{aligned}
||\tilde{\textbf{w}}_j\odot(\tilde{\textbf{e}}_j^i-\tilde{\mathbf{u}}_i v_{ji})||_{L_1} =||\textbf{e}-\mathbf{u}v_{ji}||_{L_1} \\
=\sum_{l=1}^{m} |e_l-u_l v_{ji}| = \sum_{l=1}^{m} |u_l|\cdot |v_{ji}-\frac{e_l}{u_l}|.
\end{aligned}
\end{equation}
Hence, the optimal $v_{ji}^{\ast}$ can be obtained as
\begin{equation} \label{eq:optimal_v}
\begin{aligned}
v_{ji}^{\ast} &= {\rm argmin}_{v_{ji}}||\tilde{\textbf{w}}_j\odot(\tilde{\textbf{e}}_j^i-\tilde{\mathbf{u}}_i v_{ji})||_{L_1} \\
&={\rm argmin}_{v_{ji}}\sum_{l=1}^{m} |u_l|\cdot |v_{ji}-\frac{e_l}{u_l}|.
\end{aligned}
\end{equation}
From Eq. (\ref{eq:optimal_v}), it can be seen that  $v_{ji}^{\ast}$ coincides with the weighted median of the sequence $\{\frac{e_l}{u_l}\}_{l=1}^{m}$ under weights $\{|u_l|\}_{l=1}^{m}$. By adopting the similar derivation process, $u_{ji}^{\ast}\ (j=1,\cdots,m,i=1,\cdots,r)$, the optimal value for each element $u_{ji}$ of $\mathbf{U}$, can be expressed as
\begin{equation} \label{eq:optimal_u}
u_{ji}^{\ast} = {\rm argmin}_{u_{ji}}||\textbf{w}_j\odot(\textbf{e}_j^i-\tilde{\mathbf{v}}_i^{\rm T} u_{ji})||_{L_1},
\end{equation}
where $\textbf{w}_j$ and $\textbf{e}_j^i$ represent the $j$th row of $\textbf{W}$ and $\textbf{E}_i$, respectively. In short, the optimal $\textbf{U},\textbf{V}$ can be obtained by employing CWM to repeatedly update $v_{ji} (j=1,\cdots,n; i=1,\cdots,r)$ and $u_{ji} (j=1,\cdots,m; i=1,\cdots,r)$ until the algorithm converges. To facilitate the understanding, the following Algorithm \ref{alg:cwm} lists the main steps to attain the optimal solution of Eq. (\ref{m-step3}). Note that Algorithm \ref{alg:cwm} corresponds to step 6 in Algorithm \ref{alg:moal}.

\begin{algorithm}
	\caption{Solving Eq. (\ref{m-step3}) by the CWM method.}
	\label{alg:cwm}
	\begin{algorithmic}[1]
		\REQUIRE ~~\\ 
		The observed matrix $\textbf{X}\in\mathbb{R}^{m\times n}$; the index set $\Omega$; $\pi_s,\lambda_s$ and $\gamma_{ijs}\ (i=1,\cdots,m,j=1,\cdots,n,s=1,\cdots,S)$; initial value of $\textbf{U},\textbf{V}$.
		\ENSURE ~~\\ 
		The optimal $\textbf{U},\textbf{V}$.
		\STATE Calculate each element $w_{ij}$ of $\textbf{W}$ by Eq. (\ref{eq:w_ij}). 	
		\WHILE {the convergence criterion does not satisfy}
		\STATE Cyclicly apply the weighted median filter to update each entry $v_{ji}\ (j=1,\cdots,n,i=1,\cdots,r)$ of $\textbf{V}$ with all the other elements of $\textbf{U},\textbf{V}$ fixed by solving Eq. (\ref{eq:optimal_v}).
		\STATE Cyclicly apply the weighted median filter to update each entry $u_{ji}\ (j=1,\cdots,m,i=1,\cdots,r)$ of $\textbf{U}$ with all the other elements of $\textbf{U},\textbf{V}$ fixed by solving Eq. (\ref{eq:optimal_u}).
		\ENDWHILE
	\end{algorithmic}
\end{algorithm}

\subsection{Some details of Algorithm 1}
\label{sec:details_of_A1}

\noindent\textbf{Tuning the number of components $S$ in MoAL:} Too large $S$ violates Occam Razor's principle, while too small $S$ leads to poor performance. In consequence, as described in step 8 of Algorithm \ref{alg:moal}, we employ an effective method to tune $S$. To begin with, we initialize $S$ to be a relatively small number such as $4, 5, \cdots, 8$. After each iteration, we compute the cluster that $x_{ij}$ belongs to, by $\textbf{C}(i,j)=\arg\max_s\gamma_{ijs}$. If there is no entry belonging to cluster $s$, we remove the corresponding ALD component.

\noindent\textbf{Initialization:} In Algorithm \ref{alg:moal}, the entries in $\mathbf{U}$ and $\mathbf{V}$ can be initialized by using a procedure analogous to that used in \cite{Meng2014Robust}. Particularly, the $(i,j)$th entry $u_{ij}$ of $\mathbf{U}$ was initialized in our experiments as $2\xi_{ij}c-c$, where $\xi_{ij}$ denotes a random number sampled from the standard Gaussian distribution $\mathcal{N}(0,1)$. In addition, $c=\sqrt{\bar{x}/r}$ where $\bar{x}$ is the median of all entries in $\mathbf{X}$ and $r$ indicates the rank of $\mathbf{U}$ and $\mathbf{V}$. Due to the characteristics of $\mathbf{U}$ and $\mathbf{V}$, each entry of $\mathbf{V}$ was initialized similarly. Moreover, the elements in $\bm{\lambda}, \textbf{K}$ and $\bm{\pi}$ was randomly sampled from the uniform distribution on $[0,1]$. After initializing $\pi_1,\cdots,\pi_S$, they were normalized so that their sum equals to 1.

\noindent\textbf{Convergence condition:} By following the common practice of EM algorithm, we terminate the iteration if the change of $||\textbf{U}||$ is smaller than a pre-defined value or the maximum iteration number is reached.

\section{Experimental Studies}
\label{sec:experiments}

We carried out experiments in this section to examine the performance of AQ-LRMF model. Several state-of-the-art methods were considered, including four robust LRMF methods (namely, MoG \cite{Meng2014Robust} \footnote{\url{http://www.gr.xjtu.edu.cn/c/document_library/get_file?folderId=1816179&name=DLFE-32163.rar}}, CWM \cite{CWM}, Damped Wiberg (DW) \footnote{\url{http://www.vision.is.tohoku.ac.jp/us/download/}} \cite{Okatani2011EfficientAF}, RegL1ALM \footnote{\url{https://sites.google.com/site/yinqiangzheng/}} \cite{RegL1ALM}) and a robust PCA method (SPCP solved by quasi Newton method) \footnote{\url{https://github.com/stephenbeckr/fastRPCA}} \cite{EPFL-CONF-199542}. We wrote the programming code for CWM with Matlab software. For the other compared algorithms, the codes provided by the corresponding authors were available. Since SPCP does not work in presence of missing entries, it was thus excluded from some experiments which involve missing data. Notice that DW is only considered in section \ref{sec:synthetic_exp} because it meets the ``out of memory'' problem for large-scale datasets. In the meantime, we assigned the same rank to all the considered algorithms except for SPCP since it can automatically determine the rank. To make the comparison more fair, all algorithms were initialized with the same values. Each algorithm was terminated when either 100 iterative steps are reached or the change of $||\textbf{U}||$ is less than $1\times 10^{-50}$. In order to simplify notations, our proposed method AQ-LRMF was denoted as AQ in later discussions. All the experiments were conducted with Matlab R2015b and run on a computer with Intel Core CPU 2.30 GHz, 4.00 GB RAM and Windows 7(64-bit) system.

The remainder of this section has the following structure. Section \ref{sec:synthetic_exp} studies the performance of each algorithm on synthetic data in the presence of various kinds of noise as well as missing values. Because LRMF has been applied in many fields, we also examined the performance of the compared algorithms on several real-world tasks. Sections \ref{sec:image_inpainting} and \ref{sec:multispectral_image} employ some inpainted and multispectral images to investigate how the compared algorithms behave on real images which contain missing values and various kinds of noise, respectively. Finally, sections \ref{sec:face_model} and \ref{sec:HSI} examine the performance of all algorithms on face modeling and hyperspectral image processing tasks. Table \ref{tab:info} summarizes the basic information of real-world data sets.

\begin{table*}
	\renewcommand{\arraystretch}{0.80}
	\centering
	\caption{The basic information of the used real-world data sets.}
	{\footnotesize{\begin{tabular*}{1.0\textwidth}{@{\extracolsep\fill}lllc@{\extracolsep\fill}}	
				\toprule
				Data set & \multicolumn{1}{c}{Type of task} & \multicolumn{1}{c}{Size}  & \multicolumn{1}{c}{subsection} \\
				\midrule
				CAVE     & image denoising & $262144\times31$ & 4.3 \\
				Extended Yale B & face modeling & $32256\ \ \times64$ & 4.4 \\
				Urban & hyperspectral image reconstruction & $94249\ \ \times210$ & 4.5 \\
				Terrain & hyperspectral image reconstruction & $153500\times210$ & 4.5 \\
				\bottomrule
	\end{tabular*}}}%
	\label{tab:info}%
\end{table*}%

\subsection{Synthetic experiments}
\label{sec:synthetic_exp}

First, we compared the behavior of each method with synthetic data containing different kinds of noise. Similar to \cite{Cao2015Low}, we randomly generated 30 low rank matrices $\textbf{X}=\textbf{UV}^{\rm T}$ of size $40\times20$ for each case, where $\textbf{U}\in \mathbb{R}^{40\times r}$ and $\textbf{V}\in \mathbb{R}^{20\times r}$ were sampled from the standard Gaussian distribution $\mathcal{N}(0,1)$. In particular, we considered the situations with $r=4$ and $r=8$. In the experiment, we stochastically set 20\% entries of $\textbf{X}$ as missing data and corrupted the non-missing entries with the following three groups of noise, respectively. (i) The first group include 4 kinds of heavy-tailed noise, i.e.,  $Lap(0,1.5)$ (Laplace noise with scale parameter $b=1.5$ and location parameter $\mu=0$), Gaussian noise with $\mu=0,\sigma=5$ and Student's $t$ noise with degrees of freedom 1 and 2, respectively. (ii) Two kinds of skew noise are included in the second group, i.e., asymmetric Laplace noise with $\lambda=1,\kappa=0.7$ and skew normal noise with $\sigma=3, \kappa=0.7$. (iii) Two kinds of mixture noise are included in the last group. The first one is $0.5\mathcal{N}(0,1)+0.3Lap(0,1)+0.2Lap(0,2)$ and another one is $0.5\mathcal{N}(0,1)+0.3Lap(0,1)+0.2AL(0,1,0.8)$. It is worthwhile to mention that the two mixture noises simulate the noise contained in real data, where most entries are corrupted by standard Gaussian noise and the rest entries are corrupted by heavy-tailed or skew noise. To evaluate the performance of each method, we employed the average ${L_1}$ and $L_2$ errors which are defined as $\frac{1}{mn}\sum_{i=1}^{m}\sum_{j=1}^{n}|x_{ij}-\mathbf{u}_i\mathbf{v}_j^{\rm T}|$ and $\sqrt{\frac{1}{mn}\sum_{i=1}^{m}\sum_{j=1}^{n}(x_{ij}-\mathbf{u}_i\mathbf{v}_j^{\rm T})^2}$, respectively.

\begin{figure}
	\centering
	\includegraphics[width=\linewidth]{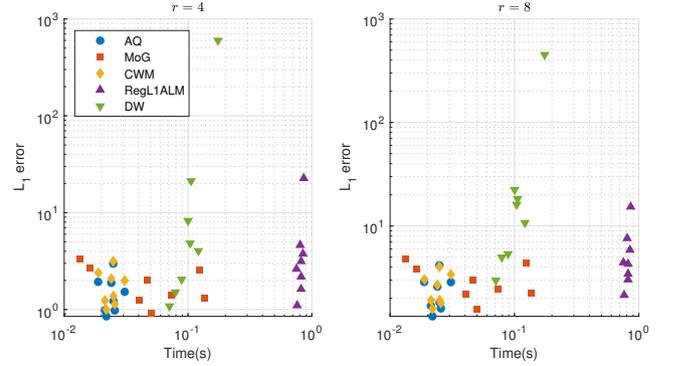}
	\caption{The scatter gram of the $L_1$ errors versus the running times of each algorithm on synthetic data.}
	\label{fig:time:fig}
\end{figure}

In our experiments, the value for the parameter $r$ in all algorithms but SPCP was set as the true rank $r$ that was used to generate synthetic data. For each compared algorithm, Tables \ref{tab:ex1-1} and \ref{tab:ex1-2} summarize the $L_1$ and $L_2$ errors averaged over 30 randomly generated matrices when $r=4$ and $r=8$, respectively. In the last two rows of Tables \ref{tab:ex1-1} and \ref{tab:ex1-2}, we list the mean and median of the $L_1$ errors as well as the $L_2$ errors of all cases. In the situation with $r=4$, it is quite obvious that our method reaches the minimum $L_1$ and $L_2$ errors for each type of noise, while MoG and CWM almost take the second place. And the approaches RegL1ALM and DW can hardly deal with the heavy-tailed and skew noise well. Note that two critical techniques are employed in AQ, that is, asymmetric noise modeling by an MoAL and solving the weighted $L_1$-norm LRMF by CWM. Based on the superiority of AQ over CWM as demonstrated in Tables \ref{tab:ex1-1} and \ref{tab:ex1-2}, we can conclude that asymmetric noise modeling indeed plays an important role in AQ for it achieving better performance. From the results corresponding to $r=8$, similar conclusions can be drawn. However, CWM evidently outperforms MoG under this circumstance, which indicates that MoG may be instable when the real rank in observed data is high. In addition, the running speed of AQ is fairly competitive, as shown in Table \ref{fig:time}. To compare the algorithms in a clearer manner, we also demonstrate two scattergrams of the $L_1$ errors versus the running times of each algorithm in Figure \ref{fig:time:fig}. It can be seen that AQ strikes a quite good balance between the reconstruction accuracy and time complexity. Although the $L_1$ errors of MoG are comparable with those of AQ and CWM, it costs more time. Moreover, CWM is observed to have almost the same time complexity with AQ, but it is outperformed by AQ in terms of reconstruction.

\begin{figure}
	\centering
	\includegraphics[width=\linewidth]{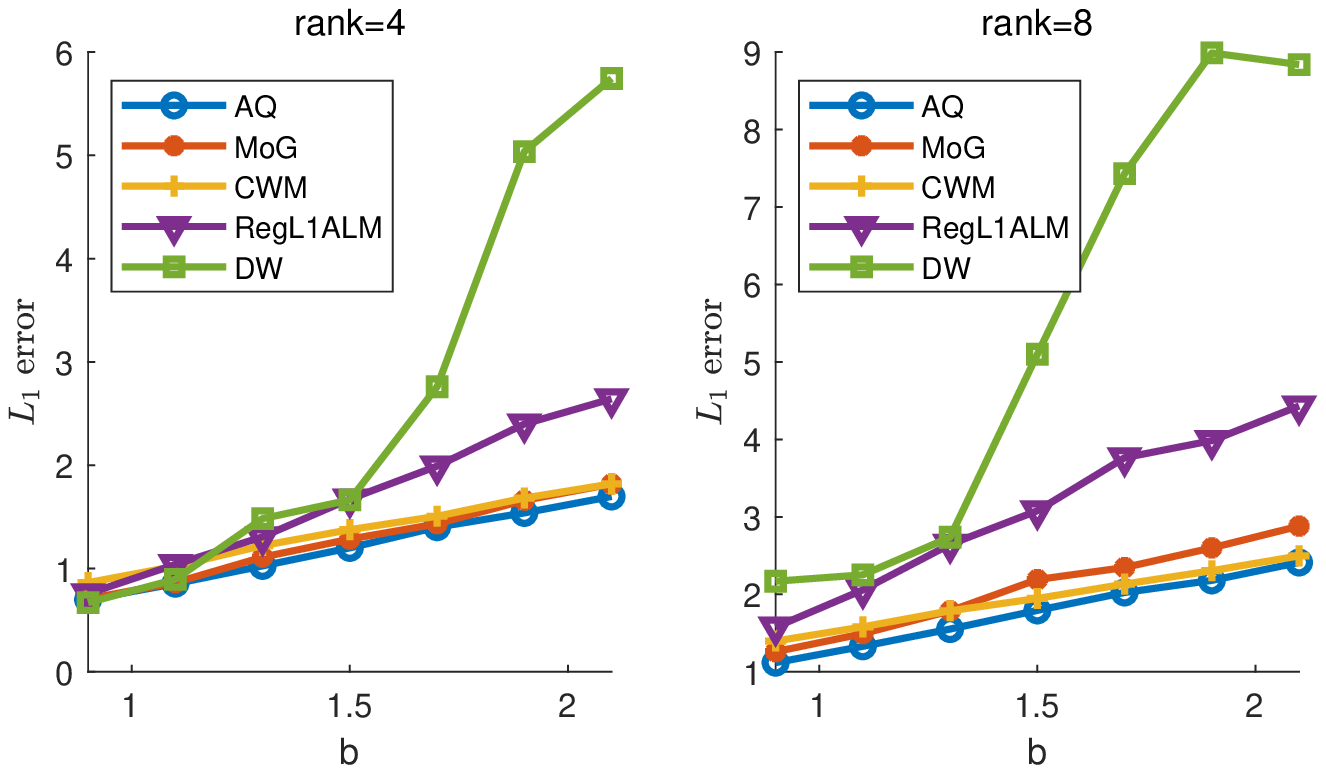}
	\caption{The performance of each algorithm on synthetic data with different levels of noise.}
	\label{fig:noiselevel}
\end{figure}

Aiming at investigating the behavior of each algorithm more extensively, we also did experiments by varying noise level under a specific type of noise. As an example, we used Laplace noise to generate data with different levels of noise. The scale parameter $b$ was varied from 0.9 to 2.1 with increment 0.2. Under each situation, the experiment was carried out similarly to the previous synthetic experiments. In Fig. \ref{fig:noiselevel}, the $L_1$ errors of each algorithm are plotted as a function of the parameter $b$. It can be observed from Fig. \ref{fig:noiselevel} that AQ almost always outperforms the other counterparts at each noise level. In summary, the simulation results presented in this subsection strongly indicate that AQ is a very competitive LRMF tool to cope with the tasks involving different kinds of noise.

\begin{table*}
	\renewcommand{\arraystretch}{0.75}
	\setlength{\tabcolsep}{3pt}
	\centering
	\caption{The average $L_1$ and $L_2$ errors for each algorithm on synthetic data with rank 4. The best and second best results are highlighted in bold and italic typeface, respectively.}
	{\footnotesize{\begin{tabular*}{1.0\textwidth}{@{\extracolsep\fill}lccccccccccc@{\extracolsep\fill}}
				\hline
				\multicolumn{1}{c}{\multirow{2}[4]{*}{$r=4$}} & \multicolumn{5}{c}{$L_1$ error}       & \multirow{2}[4]{*}{} & \multicolumn{5}{c}{$L_2$ error} \bigstrut\\
				\cline{2-6}\cline{8-12}          & AQ    & MoG   & CWM   & RegL1ALM & DW    &       & AQ    & MoG   & CWM   & RegL1ALM & DW \bigstrut\\
				\hline
				Laplace Noise (b=1.5) & \textbf{1.22 } & \textit{1.24 } & 1.38  & 1.63  & 1.51  &       & \textbf{1.82 } & \textit{1.88 } & 1.92  & 4.60  & 4.61  \bigstrut[t]\\
				Gaussian Noise ($\sigma=5$) & \textbf{2.97 } & 3.31  & \textit{3.15 } & 4.62  & 4.03  &       & \textbf{4.66 } & 5.94  & \textit{4.14 } & 13.29  & 13.94  \\
				Student's $t$ Noise ($df=1$) & \textbf{1.52 } & 2.53  & \textit{1.99 } & 22.66  & 598.22  &       & \textbf{3.41 } & 13.51  & \textit{4.79 } & 239.27  & 11952.61  \\
				Student's $t$ Noise ($df=2$) & \textbf{0.98 } & 1.32  & \textit{1.14 } & 2.18  & 8.24  &       & \textbf{1.56 } & 3.59  & \textit{1.63 } & 11.38  & 153.04  \\
				AL Noise ($\lambda=1,\kappa=0.7$) & \textbf{1.93 } & 2.68  & \textit{2.40 } & 3.76  & 4.83  &       & \textbf{2.90 } & 6.93  & \textit{3.36 } & 13.18  & 42.23  \\
				SN Noise ($\sigma=3, \kappa=0.7$) & \textbf{1.89 } & \textit{2.02 } & 2.08  & 2.62  & 2.04  &       & \textbf{2.65 } & 3.16  & \textit{2.75 } & 6.41  & 2.99  \\
				Mixture Noise 1 & \textbf{0.85 } & \textit{0.91 } & 1.00  & 1.10  & 1.08  &       & \textbf{1.23 } & 1.47  & \textit{1.43 } & 3.25  & 3.45  \\
				Mixture Noise 2 & \textbf{0.98 } & 1.40  & \textit{1.24 } & 3.15  & 21.23  &       & \textbf{1.62 } & 3.95  & \textit{1.94 } & 18.70  & 483.78  \\ \midrule
				mean  & \textbf{1.54 } & 1.93  & \textit{1.80 } & 5.22  & 80.15  &       & \textbf{2.48 } & 5.05  & \textit{2.74 } & 38.76  & 1582.08  \\
				median & \textbf{1.37 } & 1.71  & \textit{1.68 } & 2.88  & 4.43  &       & \textbf{2.24 } & 3.77  & \textit{2.35 } & 12.28  & 28.08  \bigstrut[b]\\
				\hline
	\end{tabular*}}}
	\label{tab:ex1-1}%
\end{table*}%

\begin{table*}
	\renewcommand{\arraystretch}{0.75}
	\setlength{\tabcolsep}{3pt}
	\centering
	\caption{The average $L_1$ and $L_2$ errors for each algorithm on synthetic data with rank 8. The best and second best results are highlighted in bold and italic typeface, respectively.}
	{\footnotesize{\begin{tabular*}{1.0\textwidth}{@{\extracolsep\fill}lccccccccccc@{\extracolsep\fill}}
				\toprule
				\multicolumn{1}{c}{\multirow{2}[4]{*}{$r=8$}} & \multicolumn{5}{c}{$L_1$ error}       & \multirow{2}[4]{*}{} & \multicolumn{5}{c}{$L_2$ error} \\
				\cmidrule{2-6}\cmidrule{8-12}          & AQ    & MoG   & CWM   & RegL1ALM & DW    &       & AQ    & MoG   & CWM   & RegL1ALM & DW \\
				\midrule
				Laplace Noise (b=1.5) & \textbf{1.82 } & 2.18  & \textit{1.92 } & 3.03  & 4.98  &       & \textit{2.86 } & 4.49  & \textbf{2.81 } & 8.88  & 42.57  \\
				Gaussian Noise ($\sigma=5$) & \textit{4.17 } & 4.86  & \textbf{4.04 } & 7.60  & 10.74  &       & \textit{6.18 } & 8.26  & \textbf{5.39 } & 20.64  & 82.87  \\
				Student's $t$ Noise ($df=1$) & \textbf{2.87 } & 4.36  & \textit{3.41 } & 15.33  & 450.20  &       & \textbf{8.79 } & 18.56  & \textit{11.87 } & 77.32  & 8917.25  \\
				Student's $t$ Noise ($df=2$) & \textbf{1.60 } & 2.24  & \textit{1.80 } & 3.44  & 22.36  &       & \textbf{2.61 } & 5.93  & \textit{2.82 } & 13.08  & 397.99  \\
				AL Noise ($\lambda=1,\kappa=0.7$) & \textbf{2.88 } & 3.88  & \textit{3.04 } & 5.87  & 16.09  &       & \textit{4.48 } & 9.28  & \textbf{4.35 } & 17.70  & 227.67  \\
				SN Noise ($\sigma=3, \kappa=0.7$) & \textbf{2.59 } & 3.00  & \textit{2.70 } & 4.44  & 5.37  &       & \textit{3.67 } & 5.00  & \textbf{3.66 } & 11.92  & 35.97  \\
				Mixture Noise 1 & \textbf{1.34 } & \textit{1.59 } & 1.59  & 2.15  & 2.98  &       & \textbf{2.09 } & 3.34  & \textit{2.40 } & 6.32  & 23.29  \\
				Mixture Noise 2 & \textbf{1.69 } & 2.43  & \textit{1.92 } & 4.30  & 18.33  &       & \textbf{2.88 } & 5.94  & \textit{3.09 } & 16.19  & 203.50  \\ \midrule
				mean  & \textbf{2.37 } & 3.07  & \textit{2.55 } & 5.77  & 66.38  &       & \textbf{4.19 } & 7.60  & \textit{4.55 } & 21.51  & 1241.39  \\
				median & \textbf{2.21 } & 2.72  & \textit{2.31 } & 4.37  & 13.41  &       & \textbf{3.27 } & 5.94  & \textit{3.38 } & 14.64  & 143.19  \\
				\bottomrule
	\end{tabular*}}}
	\label{tab:ex1-2}%
\end{table*}%

\begin{table*}[pos=h]
	\renewcommand{\arraystretch}{0.75}
	\setlength{\tabcolsep}{2pt}
	\centering
	\caption{The running time (in seconds) of each algorithm on synthetic data.}
	{\footnotesize{\begin{tabular*}{\textwidth}{@{\extracolsep\fill}lccccccccccc@{\extracolsep\fill}}	
				\toprule
				\multirow{2}[4]{*}{} & \multicolumn{5}{c}{$r=4$}             & \multirow{2}[4]{*}{} & \multicolumn{5}{c}{$r=8$} \\
				\cmidrule{2-6}\cmidrule{8-12}          & AQ    & MoG   & CWM   & RegL1ALM & DW    &       & AQ    & MoG   & CWM   & RegL1ALM & DW \\
				\midrule
				Laplace Noise (b=1.5) & \textbf{0.02501 } & 0.04044  & \textit{0.02502 } & 0.81899  & 0.07907  &       & \textbf{0.07898 } & 0.13455  & \textit{0.07900 } & 1.33627  & 0.26134  \\
				Gaussian Noise ($\sigma=5$) & \textit{0.02490 } & \textbf{0.01330 } & 0.02490  & 0.80570  & 0.12140  &       & \textit{0.07972 } & \textbf{0.06740 } & 0.07974  & 1.35875  & 0.42626  \\
				Student's $t$ Noise ($df=1$) & \textbf{0.03080 } & 0.12480  & \textit{0.03080 } & 0.86030  & 0.17470  &       & \textbf{0.08268 } & 0.21673  & \textit{0.08268 } & 1.26710  & 0.28241  \\
				Student's $t$ Noise ($df=2$) & \textbf{0.02560 } & 0.13690  & \textit{0.02560 } & 0.81970  & 0.10020  &       & \textbf{0.06157 } & 0.22075  & \textit{0.06157 } & 1.25522  & 0.27744  \\
				AL Noise ($\lambda=1,\kappa=0.7$) & \textit{0.01885 } & \textbf{0.01619 } & 0.01887  & 0.84787  & 0.10406  &       & \textbf{0.07640 } & 0.12882  & \textit{0.07640 } & 1.30657  & 0.29866  \\
				SN Noise ($\sigma=3, \kappa=0.7$) & \textbf{0.02399 } & 0.04659  & 0.02399  & 0.75145  & 0.08879  &       & \textit{0.07675 } & \textbf{0.07408 } & 0.07675  & 1.23501  & 0.25752  \\
				Mixture Noise 1 & \textbf{0.02189 } & 0.05013  & \textit{0.02189 } & 0.76148  & 0.07082  &       & \textbf{0.07168 } & 0.15263  & \textit{0.07169 } & 1.21608  & 0.24865  \\
				Mixture Noise 2 & \textbf{0.02133 } & 0.07331  & \textit{0.02134 } & 0.81955  & 0.10586  &       & \textbf{0.07214 } & 0.17796  & \textit{0.07214 } & 1.31874  & 0.31770  \\
				\midrule
				mean  & \textbf{0.02404 } & 0.06271  & \textit{0.02405 } & 0.81063  & 0.10561  &       & \textbf{0.07499 } & 0.14661  & \textit{0.07500 } & 1.28672  & 0.29625  \\
				median & \textbf{0.02444 } & 0.04836  & \textit{0.02444 } & 0.81927  & 0.10213  &       & \textbf{0.07657 } & 0.14359  & \textit{0.07658 } & 1.28684  & 0.27993  \\
				\bottomrule
	\end{tabular*}}}%
	\label{fig:time}
\end{table*}%

\begin{figure*}
	\centering
	\includegraphics[width=1\linewidth]{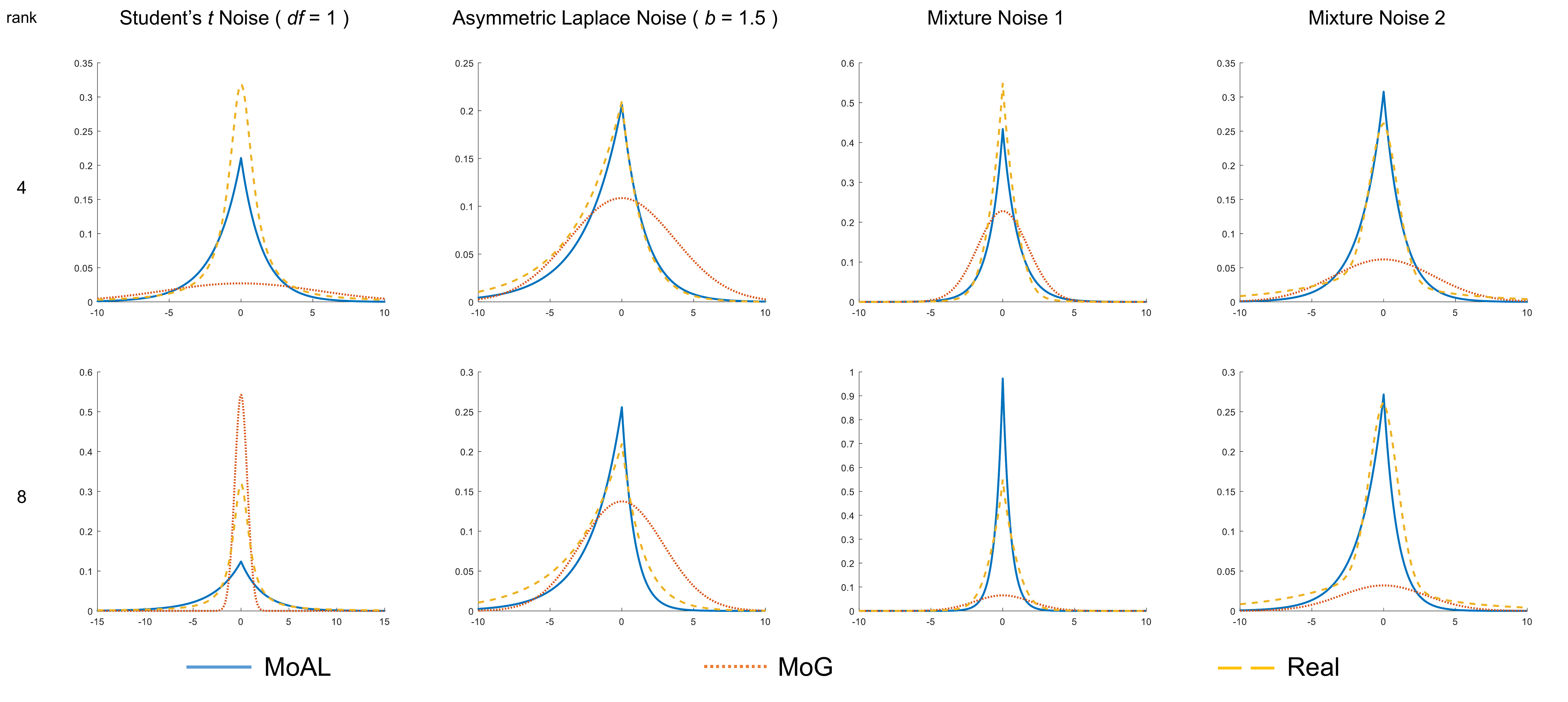}
	\caption{The comparison of the PDFs for real noise and the ones fitted by AQ and MoG in the synthetic experiments. }
	\label{fig:fig3}
\end{figure*}

To delve into the difference between AQ and MoG, we further compared the distributions of the residuals corresponding to AQ and MoG. Here, the PDFs for real nosie were utilized the groundtruth. Specifically, two symmetric and two asymmetric cases are illustrated in Figure \ref{fig:fig3}. Here, the shown PDFs fitted by AQ and MoG correspond to those reach the maximum likelihood over 30 random experiments. It is obvious that AQ does a much better job to approximate the real noise than MoG. Particularly, AQ almost provides a duplicate of real noise. In contrast, MoG is able to fit the tails, while, at the same time, it results in bad approximation to peaks. Hence, AQ has stronger power in fitting complex noise than MoG.

\subsection{Image inpainting experiments}
\label{sec:image_inpainting}

Image inpainting is a typical image processing task. In real applications, some parts of an image may be deteriorated so that the corresponding information is lost. To facilitate the understanding of the image, some sophisticated technique need to be adopted to recover its corrupted parts. This is exactly the objective of image inpainting. There is evidence that many images are low-rank matrices so that the single image inpainting can be done by matrix completion \cite{Hu2013Fast}. In image inpainting, the corrupted pixels are viewed as missing values and then the image can be recovered by an LRMF algorithm. In this paper, three typical RGB images \footnote{\url{https://sites.google.com/site/zjuyaohu/}} of size $300\times300\times3$ were employed. In our experiments, each image was reshaped to $300\times900$. By following the common practice in the research of image inpainting, we artificially corrupted the given images by putting some masks onto them. In doing so, it is convenient to examine how well each method performs to restore the original images. Here, three kinds of masks were considered, namely, random mask where 20\% pixels were stochastically removed, text masks with big and small fonts, respectively. Some evidence \cite{Hu2013Fast} has shown that the information of a single image will be lost if the rank is set to a relatively low value. Thus, the rank was set to 80 in this experiment for all algorithms.

\begin{figure}
	\centering
	\includegraphics[width=.8\linewidth]{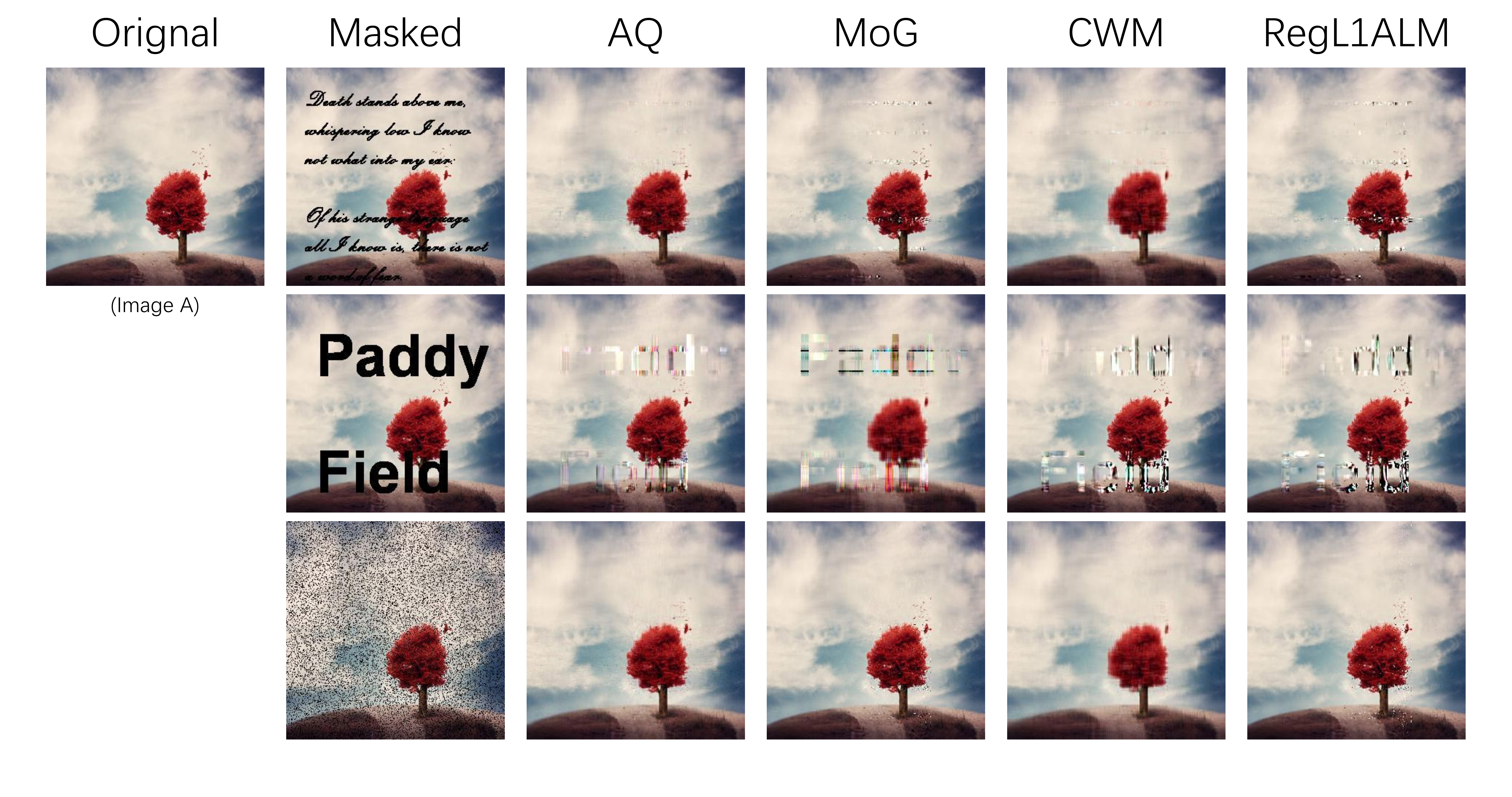}
	\includegraphics[width=.8\linewidth]{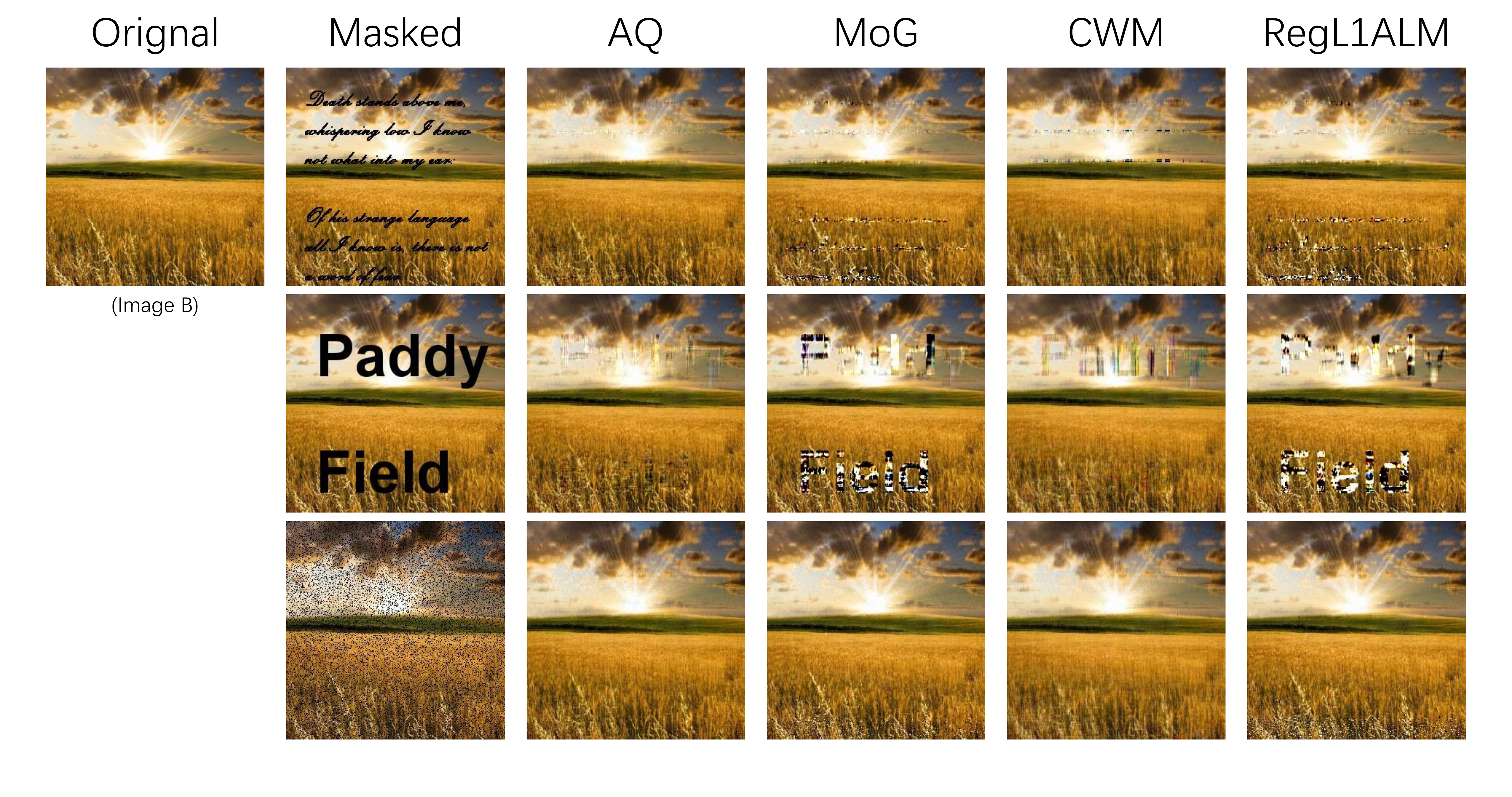}
	\includegraphics[width=.8\linewidth]{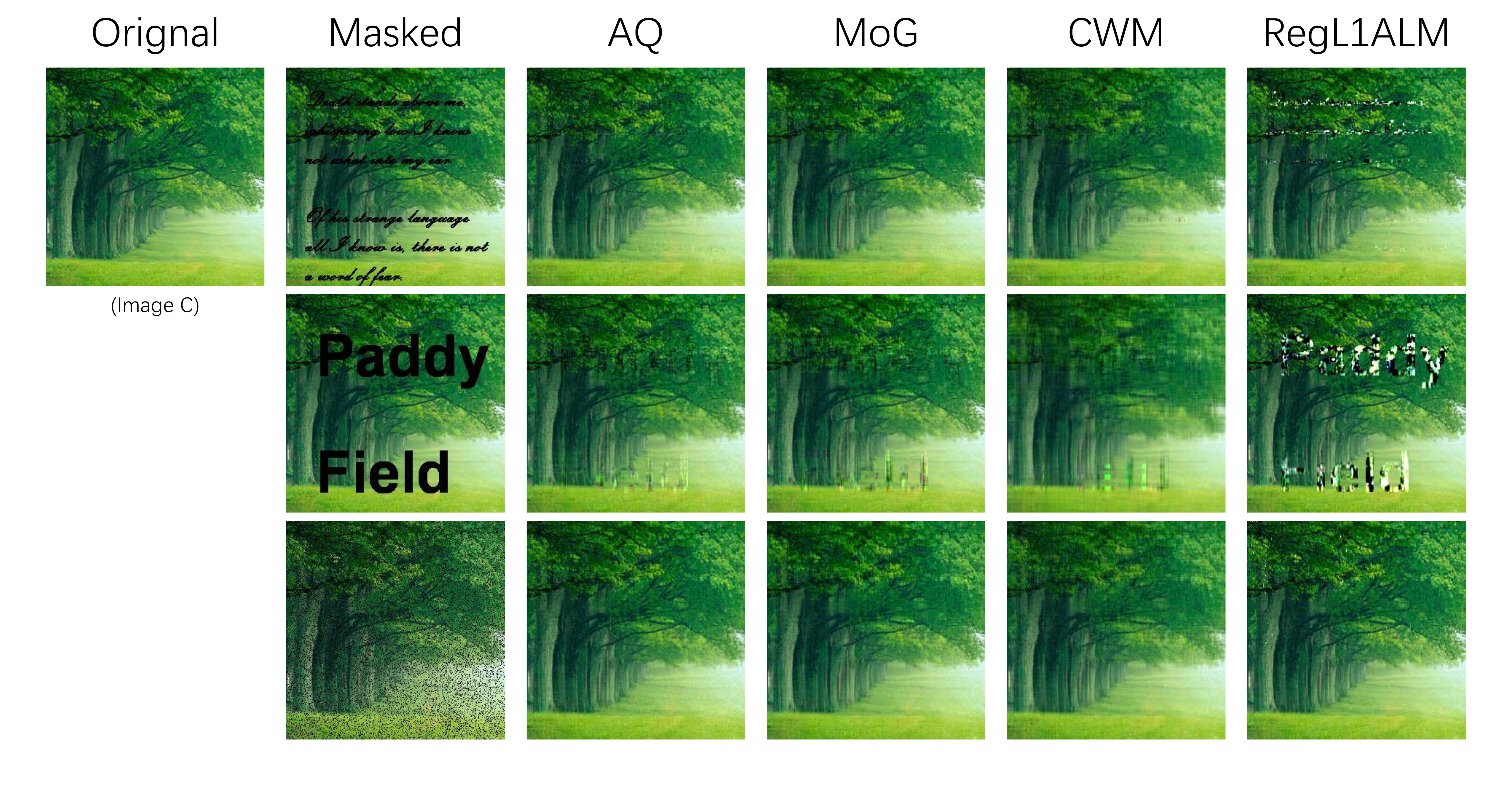}
	\caption{The original, masked and inpainting images. }
	\label{fig:fig4-1}
\end{figure}

Figure \ref{fig:fig4-1} displays the original, masked and reconstructed images, and Table \ref{tab:exp2} reports the average $L_1$ and $L_2$ errors of each algorithm. It is obvious that removing a random mask is the easiest task. In this situation, there is no significantly visible difference among the reconstructed images. AQ and MoG are the best performers. In contrast, the results shown in Figure \ref{fig:fig4-1} and Table \ref{tab:exp2} indicate that text mask removal is more difficult, especially when the images are corrupted with big fonts. The main reason lies in that the text mask is spatially correlated while it is difficult for any LRMF algorithm to effectively utilize this type of information. Under these circumstances, it can be observed in Figure \ref{fig:fig4-1} and Table \ref{tab:exp2} that AQ outperforms the other methods to remove the text masks in terms of both reconstruction error and visualization. RegL1ALM and MoG perform badly and the clear text can often be seen in their reconstructed images. Although CWM produces slightly better results, its average $L_1$ error is still higher than that of AQ. In a word, AQ possesses the superiority over the other algorithms in our investigated image inpainting tasks. In particular, AQ achieves the smallest average $L_1$ error in 5 cases and the second smallest one in 3 cases. When evaluating all algorithms with $L_2$ error, the superiority of AQ is more significant.

\begin{table*}[pos=htbp]
	\renewcommand{\arraystretch}{0.75}
	\setlength{\tabcolsep}{2pt}
	\centering
	\caption{The average $L_1$ and $L_2$ errors of each method on image inpainting experiments. The best and second best results are highlighted in bold and italic typeface, respectively.}
	{\footnotesize{\begin{tabular*}{1.0\textwidth}{@{\extracolsep\fill}clcccccccccccccc@{\extracolsep\fill}}	
				\toprule
				\multirow{2}[3]{*}{} & \multirow{2}[3]{*}{} & \multicolumn{4}{c}{Image A}   &       & \multicolumn{4}{c}{Image B}   &       & \multicolumn{4}{c}{Image C} \\
				\cmidrule{3-6}\cmidrule{8-11}\cmidrule{13-16}          &       & AQ    & MoG   & CWM   & RegL1 &       & AQ    & MoG   & CWM   & RegL1 &       & AQ    & MoG   & CWM   & RegL1 \\
				\midrule
				\multirow{3}[2]{*}{$L_1$ error} & Small & 2.59  & \textbf{2.32 } & 2.91  & \textit{2.33 } &       & \textbf{5.60 } & 7.43  & \textit{6.90 } & 8.08  &       & \textbf{5.13 } & \textit{5.83 } & 6.30  & 6.97  \\
				& Large & \textbf{5.59 } & 9.25  & \textit{7.84 } & 8.77  &       & \textbf{6.88 } & 20.16  & \textit{8.78 } & 19.11  &       & \textbf{6.84 } & \textit{7.15 } & 7.67  & 17.70  \\
				& Random & \textit{2.81 } & \textbf{2.18 } & 3.06  & 2.45  &       & \textit{5.91 } & \textbf{5.75 } & 6.65  & 6.79  &       & \textit{5.61 } & \textit{6.74 } & 6.77  & \textbf{5.07 } \\
				\midrule
				\multirow{3}[2]{*}{$L_2$ error} & Small & \textbf{5.05 } & 10.65  & \textit{6.62 } & 17.91  &       & \textbf{10.11 } & 31.13  & \textit{14.85 } & 43.26  &       & \textbf{8.67 } & \textit{9.64 } & 10.50  & 28.29  \\
				& Large & \textbf{17.95 } & 49.06  & \textit{29.23 } & 47.04  &       & \textbf{13.98 } & 79.97  & \textit{17.95 } & 75.74  &       & \textit{12.77 } & \textbf{12.65 } & 13.93  & 63.09  \\
				& Random & \textbf{5.33 } & 10.97  & \textit{6.80 } & 26.52  &       & \textbf{9.86 } & 16.90  & \textit{11.47 } & 34.06  &       & \textbf{9.18 } & \textit{10.50 } & 11.07  & 11.60  \\
				\bottomrule
	\end{tabular*}}}%
	\label{tab:exp2}
\end{table*}%

\subsection{Multispectral image experiments}
\label{sec:multispectral_image}

In this subsection, we study the behavior of all algorithms in image denoising tasks. The Columbia Multispectral Image database, CAVE, \footnote{\url{http://www1.cs.columbia.edu/CAVE/databases/multispectral}} was employed, where every scene contains 31 bands with size $512\times512$. To achieve our purpose, seven scenes out of them (i.e., Balloon, Clay, Feathers, Flowers, Hairs, Paints and Pompoms) were utilized to test the effectiveness of our methods. The used images were resized by half and the pixels were rescaled to [0,1]. Analogous to the strategy used in image inpainting experiments, some noise was artificially added to the original images. Then, each LRMF algorithm was applied to remove the noise so that the corrupted images can be restored as accurate as possible. In the experiments, three different kinds of noise were considered, that is, Laplace noise with scale parameter $b=10$, asymmetric Laplace noise with $\lambda=10,\kappa=0.7$ and mixture noise, i.e., $0.5\mathcal{N}(0,0.5)+0.3AL(0,8,0.9)+0.2AL(0,8,0.7)$. The rank was set to 4 for all algorithms.

Table \ref{tab:multispectral} reports the average $L_1$ and $L_2$ errors of each method. Evidently, AQ behaves best in most cases. For Laplace noise, RegL1ALM sometimes outperforms AQ to attain the best results. The success of RegL1ALM can be attributed to the special format of its objective function, namely, $L_1$ norm loss plus two penalties on $\mathbf{U}$ and $\mathbf{V}$. On the one hand, the $L_1$ norm loss is exactly compatible with Laplace noise. On the other hand, there is empirical evidence showing that its used penalties can lead to better performance on image datasets. For asymmetric Laplace and mixture noise, it is not surprising that AQ outperforms all the other methods. Under some circumstances, SPCP reaches the second lowest reconstruction error, while the other ones perform badly. The reason for the good behavior of SPCP may be that it does not rely on the assumption of noise distribution, while the other approaches implicitly assume that the noise distribution is not skew.

\begin{table*}[pos=htbp]
	\renewcommand{\arraystretch}{0.75}
	\setlength{\tabcolsep}{2pt}
	\centering
	\caption{The average $L_1$ and $L_2$ errors of each method on multispectral image experiments. The best and second best results are highlighted in bold and italic typeface, respectively. AL refers to asymmetric Lapalce.}
	{\footnotesize{\begin{tabular*}{1.0\textwidth}{@{\extracolsep\fill}cccccccrccccc@{\extracolsep\fill}}
				\toprule
				\multirow{2}[4]{*}{Scene} & \multirow{2}[4]{*}{Type of Noise} & \multicolumn{5}{c}{$L_1$ error}       &       & \multicolumn{5}{c}{$L_2$ error} \\
				\cmidrule{3-7}\cmidrule{9-13}          &       & AQ    & MoG   & CWM   & RegL1ALM & SPCP  &       & AQ    & MoG   & CWM   & RegL1ALM & SPCP \\
				\midrule
				\multirow{3}[2]{*}{Balloon} & Laplace & \textbf{0.0074 } & 0.0348  & 0.0398  & \textit{0.0343 } & 0.0487  &       & \textbf{0.0140 } & 0.0494  & 0.0553  & \textit{0.0487 } & 0.0693  \\
				& AL & \textbf{0.0804 } & 0.1964  & 0.1501  & 0.1379  & \textit{0.1280 } &       & \textbf{0.1053 } & 0.2310  & 0.1882  & 0.1836  & \textit{0.1510 } \\
				& Mixture & \textbf{0.1974 } & 0.2514  & 0.2453  & 0.2422  & \textit{0.2026 } &       & \textit{0.2555 } & 0.3620  & 0.3221  & 0.3879  & \textbf{0.2374 } \\
				\midrule
				\multirow{3}[2]{*}{Clay} & Laplace & \textbf{0.0335 } & 0.0402  & 0.0362  & \textit{0.0344 } & 0.0596  &       & \textbf{0.0454 } & 0.0518  & 0.0522  & \textit{0.0492 } & 0.1131  \\
				& AL & \textbf{0.0778 } & 0.1787  & 0.1385  & \textit{0.1342 } & 0.1380  &       & \textbf{0.1266 } & 0.2110  & 0.1828  & 0.1830  & \textit{0.1789 } \\
				& Mixture & \textbf{0.1569 } & 0.2169  & 0.2403  & 0.2453  & \textit{0.2097 } &       & \textbf{0.2491 } & 0.3372  & 0.3362  & 0.3714  & \textit{0.2548 } \\
				\midrule
				\multirow{3}[2]{*}{Feathers} & Laplace & 0.0392  & \textit{0.0390 } & 0.0417  & \textbf{0.0373 } & 0.0470  &       & \textbf{0.0520 } & 0.0543  & 0.0588  & \textit{0.0522 } & 0.0805  \\
				& AL & \textbf{0.0911 } & 0.1526  & 0.1487  & 0.1396  & \textit{0.1303 } &       & \textbf{0.1217 } & 0.1918  & 0.1876  & 0.1832  & \textit{0.1566 } \\
				& Mixture & \textbf{0.1946 } & 0.2575  & 0.2455  & 0.2425  & \textit{0.2046 } &       & \textit{0.2506 } & 0.3955  & 0.3272  & 0.3911  & \textbf{0.2414 } \\
				\midrule
				\multirow{3}[2]{*}{Flowers} & Laplace & \textit{0.0362 } & 0.0395  & 0.0371  & \textbf{0.0343 } & 0.0437  &       & \textbf{0.0486 } & \textit{0.0513 } & 0.0530  & 0.0526  & 0.0794  \\
				& AL & \textbf{0.0761 } & 0.1708  & 0.1450  & 0.1374  & \textit{0.1289 } &       & \textbf{0.1113 } & 0.2054  & 0.1861  & 0.1826  & \textit{0.1569 } \\
				& Mixture & \textbf{0.1709 } & 0.2339  & 0.2393  & 0.2437  & \textit{0.2025 } &       & \textit{0.2718 } & 0.3799  & 0.3307  & 0.3679  & \textbf{0.2404 } \\
				\midrule
				\multirow{3}[2]{*}{Hairs} & Laplace & 0.0321  & 0.0380  & 0.0358  & \textit{0.0292 } & \textbf{0.0253 } &       & \textit{0.0426 } & 0.0514  & 0.0511  & 0.0517  & \textbf{0.0377 } \\
				& AL & \textbf{0.0681 } & 0.1969  & 0.1373  & 0.1346  & \textit{0.1201 } &       & \textbf{0.1112 } & 0.2288  & 0.1736  & 0.1825  & \textit{0.1370 } \\
				& Mixture & \textbf{0.1412 } & 0.2172  & 0.2305  & 0.2387  & \textit{0.1979 } &       & \textit{0.2595 } & 0.3684  & 0.3288  & 0.3847  & \textbf{0.2291 } \\
				\midrule
				\multirow{3}[2]{*}{Paints} & Laplace & 0.0424  & \textit{0.0370 } & 0.0431  & \textbf{0.0354 } & 0.0396  &       & 0.0554  & \textbf{0.0497 } & 0.0648  & \textit{0.0501 } & 0.0636  \\
				& AL & \textbf{0.1009 } & 0.1910  & 0.1432  & 0.1402  & \textit{0.1259 } &       & \textbf{0.1329 } & 0.2244  & 0.1815  & 0.1831  & \textit{0.1474 } \\
				& Mixture & \textit{0.2119 } & 0.2311  & 0.2460  & 0.2399  & \textbf{0.2018 } &       & \textit{0.3070 } & 0.3777  & 0.3433  & 0.3861  & \textbf{0.2356 } \\
				\midrule
				\multirow{3}[2]{*}{Pompoms} & Laplace & 0.0493  & 0.0420  & \textit{0.0414 } & \textbf{0.0379 } & 0.0824  &       & 0.0641  & \textit{0.0542 } & 0.0548  & \textbf{0.0527 } & 0.1211  \\
				& AL & \textbf{0.1002 } & 0.1937  & 0.1508  & \textit{0.1422 } & 0.1472  &       & \textbf{0.1311 } & 0.2268  & 0.1928  & \textit{0.1847 } & 0.1850  \\
				& Mixture & \textbf{0.1630 } & 0.4121  & 0.2489  & 0.2323  & \textit{0.2153 } &       & \textbf{0.2210 } & 0.5119  & 0.3281  & 0.3887  & \textit{0.2595 } \\
				\midrule
				\multirow{3}[2]{*}{mean} & Laplace & \textbf{0.0300 } & 0.0338  & 0.0344  & \textit{0.0304 } & 0.0433  &       & \textbf{0.0403 } & 0.0453  & 0.0488  & \textit{0.0447 } & 0.0706  \\
				& AL & \textbf{0.0743 } & 0.1600  & 0.1267  & 0.1208  & \textit{0.1148 } &       & \textbf{0.1050 } & 0.1899  & 0.1616  & 0.1603  & \textit{0.1391 } \\
				& Mixture & \textbf{0.1545 } & 0.2275  & 0.2120  & 0.2106  & \textit{0.1793 } &       & \textbf{0.2268 } & 0.3416  & 0.2895  & 0.3347  & \textit{0.2123 } \\
				\bottomrule
	\end{tabular*}}}%
	\label{tab:multispectral}%
\end{table*}%

\subsection{Face modeling experiments}
\label{sec:face_model}

\begin{figure}[pos=t]
	\centering
	\includegraphics[width=1\linewidth]{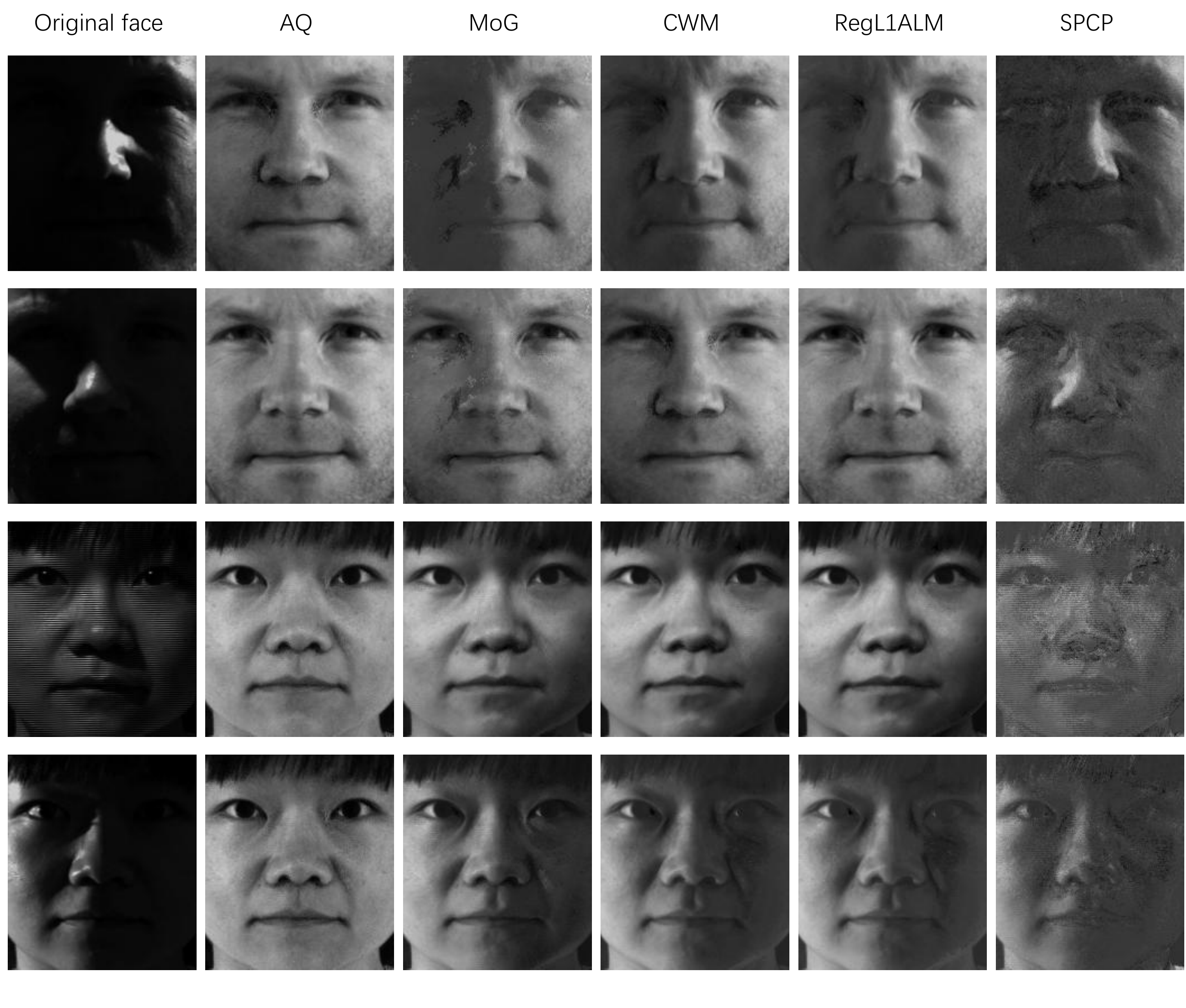}
	\caption{The original faces and the reconstructed ones.}
	\label{fig:fig4}
\end{figure}

Here, we applied the LRMF techniques to address the face modeling task. The Extended Yale B database \footnote{\url{http://cvc.yale.edu/projects/yalefaces/yalefaces.html}} consisting of 64 images with size $192\times 168$ of each subject was considered. Therefore, it leads to a $32256\times64$ matrix for each subject. Particularly, we used the face images of the third and fifth subjects. The first column of Figure \ref{fig:fig4} demonstrates some typical faces for illustration. We set the rank to 4 for all methods except for SPCP which determines the rank automatically. The second to sixth columns of Figure \ref{fig:fig4} display the faces reconstructed by the compared LRMF algorithms.

From Figure \ref{fig:fig4}, we can observe that that all methods are able to remove the cast shadows, saturations and camera noise. However, the performance of SPCP seems to be worse in comparison with other algorithms. Evidently, AQ always outperforms the other methods due to its pretty reconstruction. As shown in Figure \ref{fig:fig1}, there is an asymmetric distribution in the face with a large dark region. Because of this, the techniques MoG, CWM, RegL1ALM and SPCP which utilize the symmetric loss function lead to bad results, while AQ with the quantile loss function produces the best reconstructed images.

\subsection{Hyperspectral image experiments}
\label{sec:HSI}

\begin{figure}
	\centering
	\includegraphics[width=\linewidth]{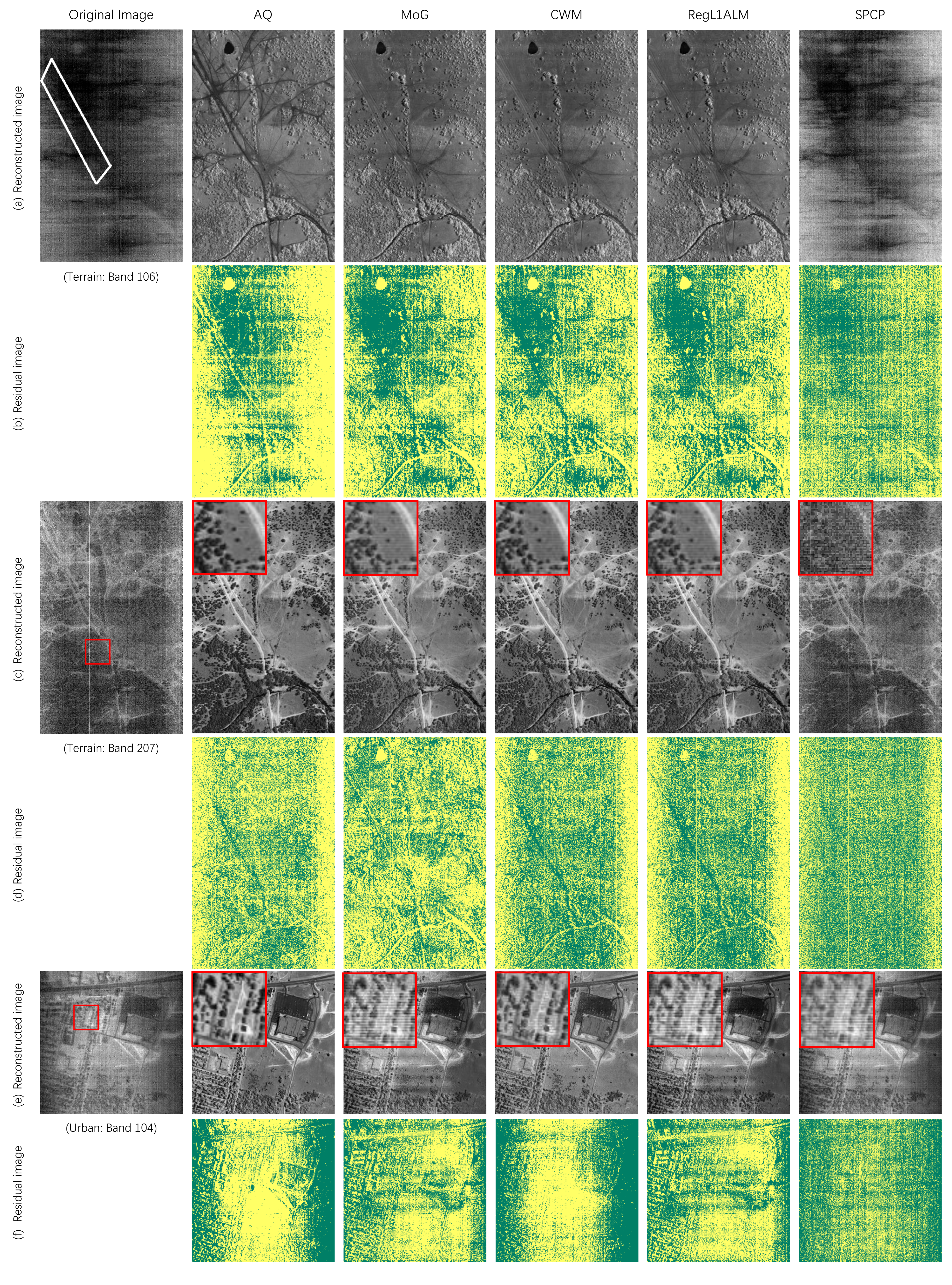}
	\caption{The reconstructed and residual images. Note that the white frame and red box are markers used to emphasize the local patch.}
	\label{fig:terrainresidual}
\end{figure}

In this subsection, we employed two HSI datasets, Urban and Terrain \footnote{\url{http://www.erdc.usace.army.mil/Media/Fact-Sheets/Fact-Sheet-Article-View/Article/610433/hypercube/}}, to investigate the behavior of all algorithms. There are 210 bands, each of which is of size $307\times 307$ for Urban and $500\times 307$ for Terrain. Thus, the data matrix is of size $94249\times 210$ for Urban and $153500\times 210$ for Terrain. Here, we utilized the same experimental settings as those used in subsection \ref{sec:face_model}. DW was still unavailable in this experiment due to the computational problem. As show in the first column of Figure \ref{fig:terrainresidual}, some parts of bands are seriously polluted by the atmosphere and water absorption.

The reconstructed images of bands 106 and 207 in the Terrain data set and the band 104 in the Urban data set are shown in Figure \ref{fig:terrainresidual} (a), (c) and (e), respectively. Their residual images (i.e., $\textbf{X}-\hat{\textbf{U}}\hat{\textbf{V}}^{\rm T}$) are also demonstrated below the reconstructed ones. Obviously, the band 106 in Terrain is seriously polluted. Nevertheless, our proposed AQ method still effectively reconstructs a clean and smooth one. Although MoG, CWM and RegL1ALM remove most parts of noise, they miss a part of local information, that is, the line from upper left corner to bottom right hand side (i.e., the white parallelogram marked in the original image). As for SPCP, it only removes few parts of noise. The residual images also reveal that AQ behaves better to deal with the detailed information. Note that the band 207 in Terrain and the band 104 in Urban are mainly corrupted by the stripe and Guassian-like noise. Under these circumstances, AQ still outperforms the others because the latter fails to remove the stripe noise. In particular, for the interested areas that are marked by rectangles and amplified areas, the bands reconstructed by MoG, CWM, RegL1ALM and SPCP contain evident stripes. As far as the reconstructed images produced by AQ are concerned, however, this phenomenon does not exist.

\begin{figure}
	\centering
	\includegraphics[width=\linewidth]{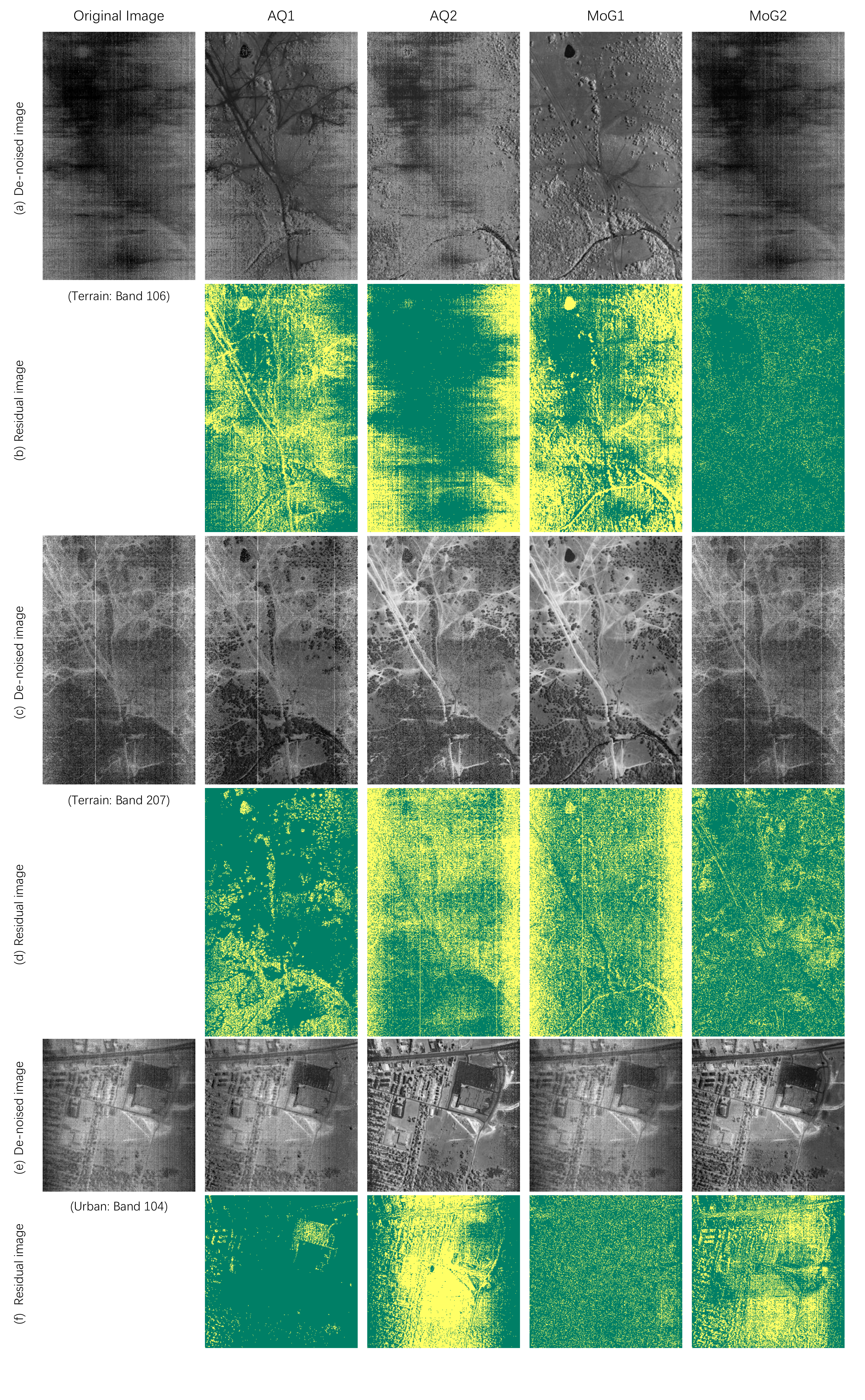}
	\caption{The de-noised and residual images produced by the two components of AQ (i.e., columns marked with AQ1 and AQ2) and MoG (i.e., columns marked with MoG1 and MoG2). For example, the image lies in the first row and second column is the de-noised image which is obtained by removing the first AQ component from the original image.}
	\label{fig:terrain2error}
\end{figure}

We conjectured that the main reason for the different behavior of these algorithms lies in their used loss function. For CWM, RegL1ALM and SPCP, too simple loss function lead them to work not well when encountering complicated noise. In contrast, AQ and MoG perform better because they use multiple distribution components to model noise. It is very interesting to study the difference between AQ and MoG. For these two algorithms, we found that they both approximate the noise in our considered three bands with two components. For AQ (MoG), we denoted them as AQ1 and AQ2 (MoG1 and MoG2), respectively. In Figure \ref{fig:terrain2error}, we presented de-noised images and residual images produced by each component. Take the de-noised image in the column AQ1 as an example, it corresponds to $\hat{\textbf{U}}\hat{\textbf{V}}^{\rm T}+{\rm AQ2}$ and the residual image shown below it corresponds to AQ1 (i.e., $\textbf{X}-\hat{\textbf{U}}\hat{\textbf{V}}^{\rm T}-{\rm AQ2})$. The other images can be understood similarly. In doing so, we can further figure out the role that each component in AQ or MoG plays. When dealing with the band 106 in Terrain, the first AQ component is seen to de-noise the center parts, while the second one targets at the left and right edges. For the band 207 in Terrain, two AQ components de-noise the bottom and the rest parts, respectively. Regarding the band 104 in Urban, they focus on the right upper and center parts, respectively. By inspecting the results generated by MoG, however, we cannot discover some regular patterns for the role that two components play. Therefore, it can be concluded that AQ can capture the local structural information of real images, although we do not encode it into our model. The reason may be that the pixels with the same skewness in real images tend to cluster. In this aspect, AQ also possesses superiority over MoG.

\section{Conclusions and future work}
\label{sec:conclusions}

Aiming at enhancing the performance of existing LRMF methods to cope with complicated noise in real applications, we propose in this work a new low-rank matrix factorization method AQ-LRMF to recover subspaces. The core idea of AQ-LRMF is to directly model unknown noise by a mixture of asymmetric Laplace distributions. We also present an efficient procedure based on the EM algorithm to estimate the parameters in AQ-LRMF. Actually, the objective function of AQ-LRMF corresponds to the adaptive quantile loss like those used in quantile regression. Nevertheless, AQ-LRMF does not need to pre-define the asymmetry parameter of quantile loss whereas quantile regression needs a user to specify its corresponding parameter in advance. Thus, AQ-LRMF has an advantage over quantile regression in this aspect. Based on the experimental results on synthetic and real data, the novel AQ-LRMF model is seen to always outperform several other state-of-the-art counterparts. In addition, AQ-LRMF also has the superiority to capture local structural information in real images. Therefore, AQ-LRMF can be deemed as a competitive tool to cope with complex real problems.

The future work are twofold. Firstly, this paper mainly investigates skew noise for images. It is very interesting to study whether skew noise is valid for other fields, such as recommender system and link prediction. Secondly, the idea of skew noise modeling can be extended to related problems, including non-negative LRMF and low-rank tensor factorization.

\section*{Statement}
The authors declare that they have no known competing interests or other things that may influence the work reported in this paper.

\section*{Acknowledgements}
The authors would like to thank the editor and the reviewers for their useful suggestions which have helped to improve the paper greatly. This research was supported by the National Key Research and Development Program of China [grant number 2018YFC0809001].

\bibliographystyle{cas-model2-names}

\bibliography{egbib.bib}

\begin{thebibliography}{40}
\expandafter\ifx\csname natexlab\endcsname\relax\def\natexlab#1{#1}\fi
\providecommand{\url}[1]{\texttt{#1}}
\providecommand{\href}[2]{#2}
\providecommand{\path}[1]{#1}
\providecommand{\DOIprefix}{doi:}
\providecommand{\ArXivprefix}{arXiv:}
\providecommand{\URLprefix}{URL: }
\providecommand{\Pubmedprefix}{pmid:}
\providecommand{\doi}[1]{\href{http://dx.doi.org/#1}{\path{#1}}}
\providecommand{\Pubmed}[1]{\href{pmid:#1}{\path{#1}}}
\providecommand{\bibinfo}[2]{#2}
\ifx\xfnm\relax \def\xfnm[#1]{\unskip,\space#1}\fi
\bibitem[{Aravkin et~al.(2914)Aravkin, Becker, Cevher and
  Olsen}]{EPFL-CONF-199542}
\bibinfo{author}{Aravkin, A.}, \bibinfo{author}{Becker, S.},
  \bibinfo{author}{Cevher, V.}, \bibinfo{author}{Olsen, P.},
  \bibinfo{year}{2914}.
\newblock \bibinfo{title}{A variational approach to stable principal component
  pursuit}, in: \bibinfo{booktitle}{Conference on Uncertainty in Artificial
  Intelligence (UAI)}, \bibinfo{address}{Arlington, Virginia, United States}.
  pp. \bibinfo{pages}{32--41}.
\bibitem[{Azzalini and Valle(1996)}]{azzalini1996the}
\bibinfo{author}{Azzalini, A.}, \bibinfo{author}{Valle, A.D.},
  \bibinfo{year}{1996}.
\newblock \bibinfo{title}{The multivariate skew-normal distribution}.
\newblock \bibinfo{journal}{Biometrika} \bibinfo{volume}{83},
  \bibinfo{pages}{715--726}.
\bibitem[{Buchanan and Fitzgibbon(2005)}]{DN}
\bibinfo{author}{Buchanan, A.M.}, \bibinfo{author}{Fitzgibbon, A.W.},
  \bibinfo{year}{2005}.
\newblock \bibinfo{title}{Damped newton algorithms for matrix factorization
  with missing data}, in: \bibinfo{booktitle}{IEEE Conference on Computer
  Vision and Pattern Recognition, San Diego, CA, USA}, pp.
  \bibinfo{pages}{316--322}.
\bibitem[{Cand\`{e}s et~al.(2011)Cand\`{e}s, Li, Ma and
  Wright}]{Candes2011Robust}
\bibinfo{author}{Cand\`{e}s, E.}, \bibinfo{author}{Li, X.},
  \bibinfo{author}{Ma, Y.}, \bibinfo{author}{Wright, J.}, \bibinfo{year}{2011}.
\newblock \bibinfo{title}{Robust principal component analysis?}
\newblock \bibinfo{journal}{Journal of the ACM} \bibinfo{volume}{58},
  \bibinfo{pages}{11}.
\bibitem[{Cao et~al.(2016)Cao, Chen, Zhao and Meng}]{Cao2015Low}
\bibinfo{author}{Cao, X.}, \bibinfo{author}{Chen, Y.}, \bibinfo{author}{Zhao,
  Q.}, \bibinfo{author}{Meng, D.}, \bibinfo{year}{2016}.
\newblock \bibinfo{title}{Low-rank matrix factorization under general mixture
  noise distributions}.
\newblock \bibinfo{journal}{IEEE Transactions on Image Processing}
  \bibinfo{volume}{25}, \bibinfo{pages}{4677--4690}.
\bibitem[{Chen(2008)}]{Chen}
\bibinfo{author}{Chen, P.}, \bibinfo{year}{2008}.
\newblock \bibinfo{title}{Optimization algorithms on subspaces: Revisiting
  missing data problem in low-rank matrix}.
\newblock \bibinfo{journal}{International Journal of Computer Vision}
  \bibinfo{volume}{80}, \bibinfo{pages}{125--142}.
\newblock \URLprefix \url{https://doi.org/10.1007/s11263-008-0135-7},
  \DOIprefix\doi{10.1007/s11263-008-0135-7}.
\bibitem[{Davino et~al.(2014)Davino, Furno and
  Vistocco}]{Davino2014Quantile_Reg}
\bibinfo{author}{Davino, C.}, \bibinfo{author}{Furno, M.},
  \bibinfo{author}{Vistocco, D.}, \bibinfo{year}{2014}.
\newblock \bibinfo{title}{Quantile regression: theory and applications}.
\newblock \bibinfo{publisher}{Hoboken: John Wiley \& Sons}.
\bibitem[{Dempster et~al.(1977)Dempster, Laird and Rubin}]{em}
\bibinfo{author}{Dempster, A.P.}, \bibinfo{author}{Laird, N.M.},
  \bibinfo{author}{Rubin, D.B.}, \bibinfo{year}{1977}.
\newblock \bibinfo{title}{Maximum likelihood from incomplete data via the em
  algorithm}.
\newblock \bibinfo{journal}{Journal of the Royal Statistical Society. Series B
  (Methodological)} \bibinfo{volume}{39}, \bibinfo{pages}{1--38}.
\bibitem[{Deng et~al.(2015)Deng, Lv, Liu, Huang, Tao and
  Gao}]{DBLP:conf/ijcai/DengLLHTG15}
\bibinfo{author}{Deng, C.}, \bibinfo{author}{Lv, Z.}, \bibinfo{author}{Liu,
  W.}, \bibinfo{author}{Huang, J.}, \bibinfo{author}{Tao, D.},
  \bibinfo{author}{Gao, X.}, \bibinfo{year}{2015}.
\newblock \bibinfo{title}{Multi-view matrix decomposition: {A} new scheme for
  exploring discriminative information}, in: \bibinfo{booktitle}{Proceedings of
  the Twenty-Fourth International Joint Conference on Artificial Intelligence,
  {IJCAI} 2015, Buenos Aires, Argentina, July 25-31, 2015}, pp.
  \bibinfo{pages}{3438--3444}.
\bibitem[{Deng et~al.(2016)Deng, Xu, Zhang, Tao, Gao and
  Li}]{DBLP:journals/tnn/DengXZTGL16}
\bibinfo{author}{Deng, C.}, \bibinfo{author}{Xu, J.}, \bibinfo{author}{Zhang,
  K.}, \bibinfo{author}{Tao, D.}, \bibinfo{author}{Gao, X.},
  \bibinfo{author}{Li, X.}, \bibinfo{year}{2016}.
\newblock \bibinfo{title}{Similarity constraints-based structured output
  regression machine: An approach to image super-resolution}.
\newblock \bibinfo{journal}{{IEEE} Trans. Neural Netw. Learning Syst.}
  \bibinfo{volume}{27}, \bibinfo{pages}{2472--2485}.
\bibitem[{Eriksson and van~den Hengel(2010)}]{L1Wiberg}
\bibinfo{author}{Eriksson, A.}, \bibinfo{author}{van~den Hengel, A.},
  \bibinfo{year}{2010}.
\newblock \bibinfo{title}{Efficient computation of robust low-rank matrix
  approximations in the presence of missing data using the $l_1$ norm}, in:
  \bibinfo{booktitle}{IEEE Conference on Computer Vision and Pattern
  Recognition (CVPR)}, pp. \bibinfo{pages}{771--778}.
\bibitem[{Fei et~al.(2017)Fei, Chen, Chong, Wang, Liu and
  He}]{Fei2017Denoising}
\bibinfo{author}{Fei, X.}, \bibinfo{author}{Chen, Y.}, \bibinfo{author}{Chong,
  P.}, \bibinfo{author}{Wang, Y.}, \bibinfo{author}{Liu, X.},
  \bibinfo{author}{He, G.}, \bibinfo{year}{2017}.
\newblock \bibinfo{title}{Denoising of hyperspectral image using low-rank
  matrix factorization}.
\newblock \bibinfo{journal}{IEEE Geoscience \& Remote Sensing Letters}
  \bibinfo{volume}{14}, \bibinfo{pages}{1141--1145}.
\bibitem[{Hu et~al.(2013)Hu, Zhang, Ye, Li and He}]{Hu2013Fast}
\bibinfo{author}{Hu, Y.}, \bibinfo{author}{Zhang, D.}, \bibinfo{author}{Ye,
  J.}, \bibinfo{author}{Li, X.}, \bibinfo{author}{He, X.},
  \bibinfo{year}{2013}.
\newblock \bibinfo{title}{Fast and accurate matrix completion via truncated
  nuclear norm regularization}.
\newblock \bibinfo{journal}{IEEE Transactions on Pattern Analysis \& Machine
  Intelligence} \bibinfo{volume}{35}, \bibinfo{pages}{2117--2130}.
\bibitem[{Ke and Kanade(2005)}]{Ke:2005:RLN:1068507.1068989}
\bibinfo{author}{Ke, Q.}, \bibinfo{author}{Kanade, T.}, \bibinfo{year}{2005}.
\newblock \bibinfo{title}{Robust $l_1$ norm factorization in the presence of
  outliers and missing data by alternative convex programming}, in:
  \bibinfo{booktitle}{IEEE Conference on Computer Vision and Pattern
  Recognition (CVPR)}, pp. \bibinfo{pages}{739--746}.
\bibitem[{Kim et~al.(2015)Kim, Lee, Choi, Kwak and Oh}]{Kim2015Efficient}
\bibinfo{author}{Kim, E.}, \bibinfo{author}{Lee, M.}, \bibinfo{author}{Choi,
  C.H.}, \bibinfo{author}{Kwak, N.}, \bibinfo{author}{Oh, S.},
  \bibinfo{year}{2015}.
\newblock \bibinfo{title}{Efficient $l_{1}$ -norm-based low-rank matrix
  approximations for large-scale problems using alternating rectified gradient
  method}.
\newblock \bibinfo{journal}{IEEE Transactions on Neural Networks and Learning
  Systems} \bibinfo{volume}{26}, \bibinfo{pages}{237--251}.
\bibitem[{Koltchinskii et~al.(2011)Koltchinskii, Lounici and
  Tsybakov}]{Koltchinskii2011NUCLEAR}
\bibinfo{author}{Koltchinskii, V.}, \bibinfo{author}{Lounici, K.},
  \bibinfo{author}{Tsybakov, A.B.}, \bibinfo{year}{2011}.
\newblock \bibinfo{title}{Nuclear-norm penalization and optimal rates for noisy
  low-rank matrix completion}.
\newblock \bibinfo{journal}{Annals of Statistics} \bibinfo{volume}{39},
  \bibinfo{pages}{2302--2329}.
\bibitem[{Kozumi and Kobayashi(2011)}]{Kozumi2011Gibbs}
\bibinfo{author}{Kozumi, H.}, \bibinfo{author}{Kobayashi, G.},
  \bibinfo{year}{2011}.
\newblock \bibinfo{title}{Gibbs sampling methods for bayesian quantile
  regression}.
\newblock \bibinfo{journal}{Journal of Statistical Computation \& Simulation}
  \bibinfo{volume}{81}, \bibinfo{pages}{1565--1578}.
\bibitem[{Lakshminarayanan et~al.(2011)Lakshminarayanan, Bouchard and
  Archambeau}]{Lakshminarayanan2011Robust}
\bibinfo{author}{Lakshminarayanan, B.}, \bibinfo{author}{Bouchard, G.},
  \bibinfo{author}{Archambeau, C.}, \bibinfo{year}{2011}.
\newblock \bibinfo{title}{Robust bayesian matrix factorisation}.
\newblock \bibinfo{journal}{Journal of Machine Learning Research}
  \bibinfo{volume}{15}, \bibinfo{pages}{425--433}.
\bibitem[{Lee and Seung(1999)}]{Lee1999Learning}
\bibinfo{author}{Lee, D.D.}, \bibinfo{author}{Seung, H.S.},
  \bibinfo{year}{1999}.
\newblock \bibinfo{title}{Learning the parts of objects by non-negative matrix
  factorization}.
\newblock \bibinfo{journal}{Nature} \bibinfo{volume}{401},
  \bibinfo{pages}{788--791}.
\bibitem[{Li et~al.(2017)Li, Zhang and Guo}]{Li2017Efficient}
\bibinfo{author}{Li, S.}, \bibinfo{author}{Zhang, J.}, \bibinfo{author}{Guo,
  X.}, \bibinfo{year}{2017}.
\newblock \bibinfo{title}{Efficient low rank matrix approximation via
  orthogonality pursuit and $l_2$ regularization}, in: \bibinfo{booktitle}{IEEE
  International Conference on Multimedia and Expo}, pp.
  \bibinfo{pages}{871--876}.
\bibitem[{Lin et~al.(2018)Lin, Xu and Zha}]{Lin2017Robust}
\bibinfo{author}{Lin, Z.}, \bibinfo{author}{Xu, C.}, \bibinfo{author}{Zha, H.},
  \bibinfo{year}{2018}.
\newblock \bibinfo{title}{Robust matrix factorization by majorization
  minimization}.
\newblock \bibinfo{journal}{IEEE Transactions on Pattern Analysis \& Machine
  Intelligence} \bibinfo{volume}{40}, \bibinfo{pages}{208--220}.
\bibitem[{Maz'Ya and Schmidt(1996)}]{Maz1996On}
\bibinfo{author}{Maz'Ya, V.}, \bibinfo{author}{Schmidt, G.},
  \bibinfo{year}{1996}.
\newblock \bibinfo{title}{On approximate approximations using gaussian
  kernels}.
\newblock \bibinfo{journal}{IMA Journal of Numerical Analysis}
  \bibinfo{volume}{16}, \bibinfo{pages}{13--29}.
\bibitem[{Meng and Torre(2014)}]{Meng2014Robust}
\bibinfo{author}{Meng, D.}, \bibinfo{author}{Torre, F.D.L.},
  \bibinfo{year}{2014}.
\newblock \bibinfo{title}{Robust matrix factorization with unknown noise}, in:
  \bibinfo{booktitle}{IEEE International Conference on Computer Vision (ICCV)},
  pp. \bibinfo{pages}{1337--1344}.
\bibitem[{Meng et~al.(2013)Meng, Xu, Zhang and Zhao}]{CWM}
\bibinfo{author}{Meng, D.}, \bibinfo{author}{Xu, Z.}, \bibinfo{author}{Zhang,
  L.}, \bibinfo{author}{Zhao, J.}, \bibinfo{year}{2013}.
\newblock \bibinfo{title}{A cyclic weighted median method for $l_1$ low-rank
  matrix factorization with missing entries}, in:
  \bibinfo{booktitle}{Proceedings of the Twenty-Seventh AAAI Conference on
  Artificial Intelligence}, pp. \bibinfo{pages}{704--710}.
\bibitem[{Okatani et~al.(2011)Okatani, Yoshida and
  Deguchi}]{Okatani2011EfficientAF}
\bibinfo{author}{Okatani, T.}, \bibinfo{author}{Yoshida, T.},
  \bibinfo{author}{Deguchi, K.}, \bibinfo{year}{2011}.
\newblock \bibinfo{title}{Efficient algorithm for low-rank matrix factorization
  with missing components and performance comparison of latest algorithms}, in:
  \bibinfo{booktitle}{IEEE International Conference on Computer Vision (ICCV)},
  pp. \bibinfo{pages}{842--849}.
\bibitem[{Okutomi et~al.(2012)Okutomi, Yan, Sugimoto, Liu and Zheng}]{RegL1ALM}
\bibinfo{author}{Okutomi, M.}, \bibinfo{author}{Yan, S.},
  \bibinfo{author}{Sugimoto, S.}, \bibinfo{author}{Liu, G.},
  \bibinfo{author}{Zheng, Y.}, \bibinfo{year}{2012}.
\newblock \bibinfo{title}{Practical low-rank matrix approximation under robust
  $l_1$-norm}, in: \bibinfo{booktitle}{IEEE Conference on Computer Vision and
  Pattern Recognition (CVPR)}, \bibinfo{address}{Los Alamitos, CA, USA}. pp.
  \bibinfo{pages}{1410--1417}.
\bibitem[{Srebro and Jaakkola(2003)}]{icml2003SrebroJ03}
\bibinfo{author}{Srebro, N.}, \bibinfo{author}{Jaakkola, T.},
  \bibinfo{year}{2003}.
\newblock \bibinfo{title}{Weighted low-rank approximations}, in:
  \bibinfo{booktitle}{International Conference on Machine Learning (ICML)}, pp.
  \bibinfo{pages}{720--727}.
\bibitem[{Toli\'{c} et~al.(2018)Toli\'{c}, Antulov-Fantulin and
  Kopriva}]{TOLIC201840}
\bibinfo{author}{Toli\'{c}, D.}, \bibinfo{author}{Antulov-Fantulin, N.},
  \bibinfo{author}{Kopriva, I.}, \bibinfo{year}{2018}.
\newblock \bibinfo{title}{A nonlinear orthogonal non-negative matrix
  factorization approach to subspace clustering}.
\newblock \bibinfo{journal}{Pattern Recognition} \bibinfo{volume}{82},
  \bibinfo{pages}{40 -- 55}.
\bibitem[{Udell et~al.(2016)Udell, Horn, Zadeh and Boyd}]{Udell2014Generalized}
\bibinfo{author}{Udell, M.}, \bibinfo{author}{Horn, C.},
  \bibinfo{author}{Zadeh, R.}, \bibinfo{author}{Boyd, S.},
  \bibinfo{year}{2016}.
\newblock \bibinfo{title}{Generalized low rank models}.
\newblock \bibinfo{journal}{Foundations and Trends in Machine Learning}
  \bibinfo{volume}{9}, \bibinfo{pages}{1--118}.
\bibitem[{Wang et~al.(2012)Wang, Yao, Wang and Yeung}]{Wang2012}
\bibinfo{author}{Wang, N.}, \bibinfo{author}{Yao, T.}, \bibinfo{author}{Wang,
  J.}, \bibinfo{author}{Yeung, D.Y.}, \bibinfo{year}{2012}.
\newblock \bibinfo{title}{A probabilistic approach to robust matrix
  factorization}, in: \bibinfo{booktitle}{European Conference on Computer
  Vision (ECCV)}, \bibinfo{organization}{Springer}. pp.
  \bibinfo{pages}{126--139}.
\bibitem[{Wang et~al.(2017a)Wang, Yang, Chen, Song, Zhang and
  Wang}]{wang2017transcriptomic}
\bibinfo{author}{Wang, P.}, \bibinfo{author}{Yang, C.}, \bibinfo{author}{Chen,
  H.}, \bibinfo{author}{Song, C.}, \bibinfo{author}{Zhang, X.},
  \bibinfo{author}{Wang, D.}, \bibinfo{year}{2017}a.
\newblock \bibinfo{title}{Transcriptomic basis for drought-resistance in
  brassica napus l.}
\newblock \bibinfo{journal}{Scientific reports} \bibinfo{volume}{7},
  \bibinfo{pages}{40532}.
\bibitem[{Wang et~al.(2017b)Wang, Feng, Jiao and Yu}]{WANG2017104}
\bibinfo{author}{Wang, W.}, \bibinfo{author}{Feng, Y.}, \bibinfo{author}{Jiao,
  P.}, \bibinfo{author}{Yu, W.}, \bibinfo{year}{2017}b.
\newblock \bibinfo{title}{Kernel framework based on non-negative matrix
  factorization for networks reconstruction and link prediction}.
\newblock \bibinfo{journal}{Knowledge-Based Systems} \bibinfo{volume}{137},
  \bibinfo{pages}{104 -- 114}.
\bibitem[{Wang et~al.(2018)Wang, Yang, Chen, Wang, Wang, Song, Zhang and
  Wang}]{WANG2018410}
\bibinfo{author}{Wang, Z.}, \bibinfo{author}{Yang, C.}, \bibinfo{author}{Chen,
  H.}, \bibinfo{author}{Wang, P.}, \bibinfo{author}{Wang, P.},
  \bibinfo{author}{Song, C.}, \bibinfo{author}{Zhang, X.},
  \bibinfo{author}{Wang, D.}, \bibinfo{year}{2018}.
\newblock \bibinfo{title}{Multi-gene co-expression can improve comprehensive
  resistance to multiple abiotic stresses in brassica napus l.}
\newblock \bibinfo{journal}{Plant Science} \bibinfo{volume}{274},
  \bibinfo{pages}{410 -- 419}.
\newblock \URLprefix
  \url{http://www.sciencedirect.com/science/article/pii/S0168945218305053},
  \DOIprefix\doi{https://doi.org/10.1016/j.plantsci.2018.06.014}.
\bibitem[{Wu et~al.(2018)Wu, Zhang, Yue, Zhang, He and Sun}]{WU201846}
\bibinfo{author}{Wu, H.}, \bibinfo{author}{Zhang, Z.}, \bibinfo{author}{Yue,
  K.}, \bibinfo{author}{Zhang, B.}, \bibinfo{author}{He, J.},
  \bibinfo{author}{Sun, L.}, \bibinfo{year}{2018}.
\newblock \bibinfo{title}{Dual-regularized matrix factorization with deep
  neural networks for recommender systems}.
\newblock \bibinfo{journal}{Knowledge-Based Systems} \bibinfo{volume}{145},
  \bibinfo{pages}{46 -- 58}.
\bibitem[{Xiong and Kong(2019)}]{XIONG2019464}
\bibinfo{author}{Xiong, H.}, \bibinfo{author}{Kong, D.}, \bibinfo{year}{2019}.
\newblock \bibinfo{title}{Elastic nonnegative matrix factorization}.
\newblock \bibinfo{journal}{Pattern Recognition} \bibinfo{volume}{90},
  \bibinfo{pages}{464 -- 475}.
\bibitem[{Xu et~al.(2017)Xu, Deng, Gao, Shen and
  Huang}]{DBLP:conf/ijcai/XuDGSH17}
\bibinfo{author}{Xu, J.}, \bibinfo{author}{Deng, C.}, \bibinfo{author}{Gao,
  X.}, \bibinfo{author}{Shen, D.}, \bibinfo{author}{Huang, H.},
  \bibinfo{year}{2017}.
\newblock \bibinfo{title}{Predicting alzheimer's disease cognitive assessment
  via robust low-rank structured sparse model}, in:
  \bibinfo{booktitle}{Proceedings of the Twenty-Sixth International Joint
  Conference on Artificial Intelligence, {IJCAI} 2017, Melbourne, Australia,
  August 19-25, 2017}, pp. \bibinfo{pages}{3880--3886}.
\bibitem[{Yang et~al.(2019)Yang, Deng and Nie}]{DBLP:journals/pr/YangDN19}
\bibinfo{author}{Yang, M.}, \bibinfo{author}{Deng, C.}, \bibinfo{author}{Nie,
  F.}, \bibinfo{year}{2019}.
\newblock \bibinfo{title}{Adaptive-weighting discriminative regression for
  multi-view classification}.
\newblock \bibinfo{journal}{Pattern Recognition} \bibinfo{volume}{88},
  \bibinfo{pages}{236--245}.
\bibitem[{Ye(2005)}]{Ye2005Generalized}
\bibinfo{author}{Ye, J.}, \bibinfo{year}{2005}.
\newblock \bibinfo{title}{Generalized low rank approximations of matrices}.
\newblock \bibinfo{journal}{Machine Learning} \bibinfo{volume}{61},
  \bibinfo{pages}{167--191}.
\bibitem[{Yu and Zhang(2005)}]{Keming2005A}
\bibinfo{author}{Yu, K.}, \bibinfo{author}{Zhang, J.}, \bibinfo{year}{2005}.
\newblock \bibinfo{title}{A three-parameter asymmetric laplace distribution and
  its extension}.
\newblock \bibinfo{journal}{Communications in Statistics - Theory and Methods}
  \bibinfo{volume}{34}, \bibinfo{pages}{1867--1879}.
\bibitem[{Zhou et~al.(2010)Zhou, Li, Wright and Cand\`{e}s}]{Zhou2010Stable}
\bibinfo{author}{Zhou, Z.}, \bibinfo{author}{Li, X.}, \bibinfo{author}{Wright,
  J.}, \bibinfo{author}{Cand\`{e}s, E.}, \bibinfo{year}{2010}.
\newblock \bibinfo{title}{Stable principal component pursuit}, in:
  \bibinfo{booktitle}{IEEE International Symposium on Information Theory
  Proceedings, Austin, TX, USA}, pp. \bibinfo{pages}{1518--1522}.

\end{thebibliography}


\bio{SXW}
Shuang Xu is currently pursing the Ph.D. degree in statistics with the School of Mathematics and Statistics, Xi'an Jiaotong University, Xi'an, China. His current research interests include Bayesian statistics, deep learning and complex network. 
\newline
\newline
\newline
\newline
\endbio

\bio{CXZW}
Chunxia Zhang received her Ph.D degree in Applied Mathematics from Xi'an Jiaotong University, Xi'an, China, in 2010.	
Currently, she is an associate professor in School of Mathematics and Statistics at Xi'an Jiaotong University. She has authored and coauthored about 30 journal papers on ensemble learning techniques, nonparametric regression and etc. Her main interests are in the area of ensemble learning, variable selection and deep learning.
\endbio

\bio{JSZ}
Jiangshe Zhang was born in 1962. He received the M.S. and Ph.D. degrees in applied mathematics from Xi'an Jiaotong University, Xi'an, China, in 1987 and 1993, respectively, where he is currently a Professor with the Department of Statistics. He has authored and co-authored one monograph and over 80 conference and journal publications on robust clustering, optimization, short-term load forecasting for electric power system, and remote sensing image processing. His current research interests include Bayesian statistics, global optimization, ensemble learning, and deep learning.
\endbio

\end{document}